%% file: pmlr.tex
\documentclass[pmlr]{jmlr}

\usepackage{longtable}

\usepackage{booktabs}
\usepackage[load-configurations=version-1]{siunitx} 

\usepackage{xcolor}
\usepackage{xspace}
\usepackage{comment}

\newcommand{\fs}{\textsc{FActScore}\xspace}
\newcommand{\subjectLM}{LM$_\textsc{subj}$\xspace}
\newcommand{\evalLM}{LM$_\textsc{eval}$\xspace}
\newcommand{\multi}{(\texttt{lang}, \texttt{lang})\xspace}
\newcommand{\trans}{(\texttt{lang}, \texttt{en})\xspace}
\newcommand{\enprompt}{(\texttt{en}, \texttt{en})\xspace}
\usepackage{algorithm}
\usepackage{algorithmic}
\usepackage{booktabs}
\usepackage{threeparttable}
\usepackage{multirow}
\usepackage{tabulary}
\usepackage{makecell}
\usepackage{amsmath} 
\usepackage{dsfont}
\usepackage{tikz}
\usepackage{pgfplots}
\usepackage{adjustbox}
\usepackage{hyperref}
\usepackage{enumitem}
\definecolor{aliceblue}{HTML}{E7EFFA}
\definecolor{cornsilk}{HTML}{FFF2CC}
\definecolor{champagnepink}{HTML}{FCE4D6}
\definecolor{honeydew}{HTML}{E2F0D9}
\definecolor{lavender}{HTML}{E6E6FA}

\definecolor{darkslategray76}{RGB}{76,76,76}
\definecolor{cadetblue74172155}{RGB}{74,172,155}
\definecolor{cadetblue75170172}{RGB}{75,170,172}
\definecolor{cadetblue80171194}{RGB}{80,171,194}
\definecolor{cornflowerblue108173226}{RGB}{108,173,226}
\definecolor{darkgray176}{RGB}{176,176,176}
\definecolor{darksalmon228148112}{RGB}{228,148,112}
\definecolor{darksalmon233149163}{RGB}{233,149,163}
\definecolor{darkslategray66}{RGB}{66,66,66}
\definecolor{lightsteelblue173171235}{RGB}{173,171,235}
\definecolor{mediumseagreen73174130}{RGB}{73,174,130}
\definecolor{olivedrab9517471}{RGB}{95,174,71}
\definecolor{peru17115771}{RGB}{171,157,71}
\definecolor{peru19815275}{RGB}{198,152,75}
\definecolor{plum206155232}{RGB}{206,155,232}
\definecolor{plum232145191}{RGB}{232,145,191}
\definecolor{violet230139219}{RGB}{230,139,219}
\definecolor{yellowgreen14416469}{RGB}{144,164,69}

\usepackage{graphicx}
\usepackage{wrapfig}
\graphicspath{ {./figures/} }
\usepackage{subcaption}

\theorembodyfont{\upshape}
\theoremheaderfont{\scshape}
\theorempostheader{:}
\theoremsep{\newline}

\jmlrvolume{1}
\jmlryear{2025}
\jmlrworkshop{Algorithmic Fairness Through the Lens of Metrics and Evaluation}

\title[Multilingual hallucination gaps]{Multilingual hallucination gaps}

\author{\Name{Cléa Chataigner} \Email{clea.chataigner@mila.quebec}\\
\addr Mila, Quebec AI Institute, 6666 Saint-Urbain Street, Montreal, QC, H2S 3H1 \\
    \addr McGill University, 845 Rue Sherbrooke O, Montréal, QC H3A 0G4
\AND
\Name{Afaf Taïk} \Email{afaf.taik@mila.quebec} \\
    \addr Mila, Quebec AI Institute, 6666 Saint-Urbain Street, Montreal, QC, H2S 3H1 \\
    \addr University of Montréal, 2900 Edouard Montpetit Blvd, Montreal, Quebec H3T 1J4
  \AND  
  \Name{Golnoosh Farnadi} \Email{farnadig@mila.quebec} \\
\addr Mila, Quebec AI Institute, 6666 Saint-Urbain Street, Montreal, QC, H2S 3H1\\
    \addr McGill University, 845 Rue Sherbrooke O, Montréal, QC H3A 0G4}

\editors{Miriam Rateike, Awa Dieng, Jamelle Watson-Daniels, Ferdinando Fioretto, Golnoosh Farnadi}

\begin{document}

\maketitle

\begin{abstract}
Large language models (LLMs) are increasingly used as alternatives to traditional search engines given their capacity to generate text that resembles human language. However, this shift is concerning, as LLMs often generate hallucinations—misleading or false information that appears highly credible.
In this study, we explore the phenomenon of hallucinations across multiple languages in free-form text generation, focusing on what we call \textit{multilingual hallucination gaps}. These gaps reflect differences in the frequency of hallucinated answers depending on the prompt and language used. To quantify such hallucinations, we used the \fs metric and extended its framework to a multilingual setting. We conducted experiments using LLMs from the LLaMA, Qwen, and Aya families, generating biographies in 19 languages and comparing the results to Wikipedia pages. Our results reveal variations in hallucination rates, especially between high- and low-resource languages, raising important questions about LLM multilingual performance and the challenges in evaluating hallucinations in multilingual free-form text generation.
\end{abstract}

\section{Introduction}
\label{sec:intro}

\begin{wrapfigure}{R}{0.4\textwidth}
\centering
\includegraphics[width=0.4\textwidth]{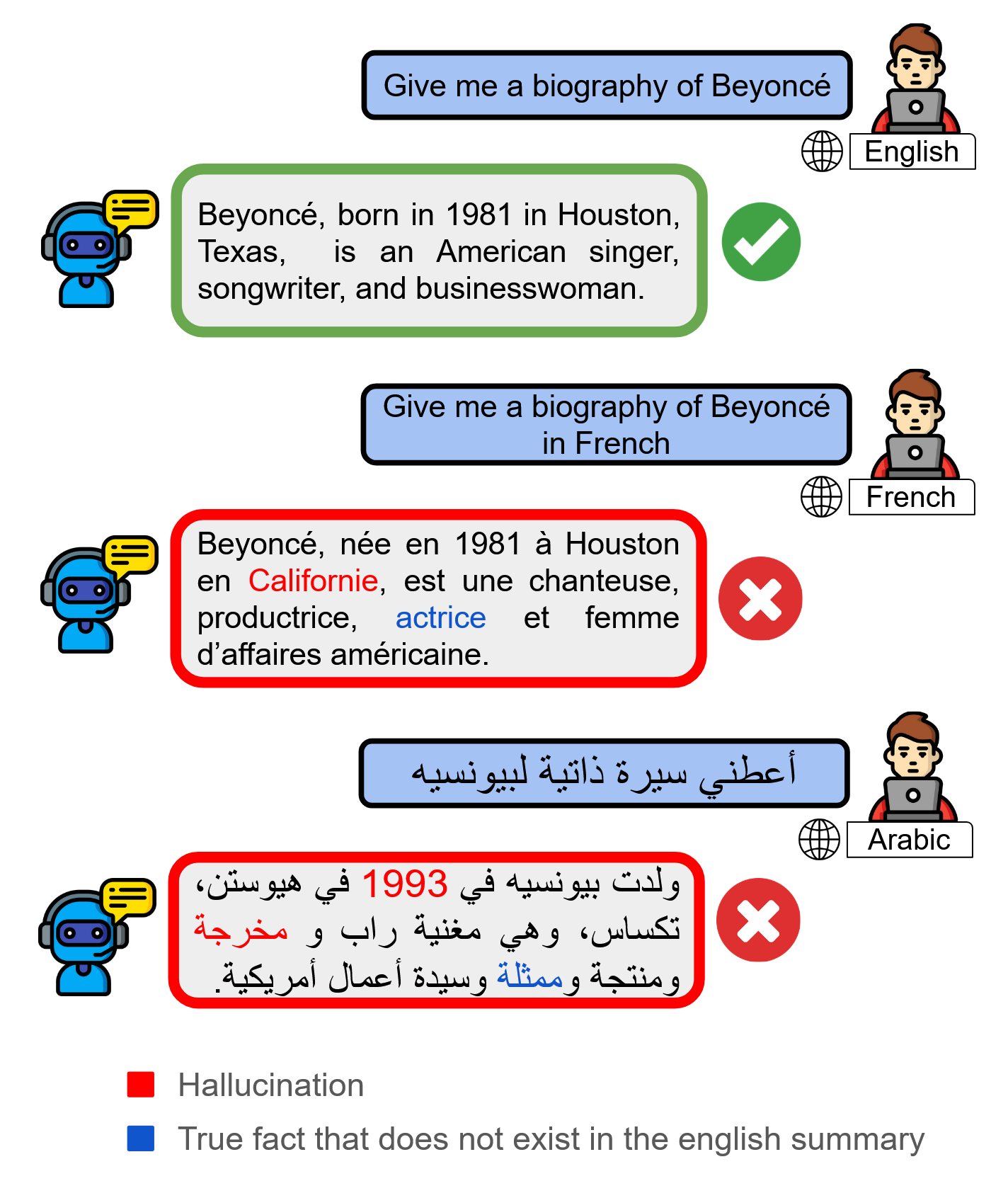}
\caption{\centering Example of Factual Hallucinations Gaps between languages}
\label{fig:intro}
\end{wrapfigure}

Since the public release of ChatGPT, large language models (LLMs) have gained popularity. They are increasingly being integrated into or even replacing traditional search engines, such as the LLaMA model for Meta mobile applications or Gemma for Google. This trend shows an increasing reliance on LLMs as sources of knowledge, due to their ability to generate human-like text. However, such use is concerning as LLMs tend to produce hallucinations.

A hallucination occurs when a LLM generates \textit{false} content \citep{rawte2023survey} with respect to a specific \textit{reference}. Based on the reference type, hallucinations can be classified as follows \citep{zhang2023sirens}: input-conflicting, where the generated content contradicts the user's input; context-conflicting, where it contradicts earlier outputs from the model; and fact-conflicting, where it contradicts established external knowledge. This work focuses exclusively on fact-conflicting hallucinations.

Understanding and tackling the issue of hallucinations in LLMs bring unique challenges. 
For example, detecting hallucinations is inherently difficult as they often appear highly credible.
The wide range of tasks that LLMs are applied to also adds to the complexity, making it harder to comprehensively evaluate and mitigate hallucinations across different applications \citep{zhang2023sirens}. 
Besides these well-known and investigated issues, hallucinations are also not produced in the same way depending on the prompt fed to the model. We introduce the concept of \textit{multilingual hallucination gaps}, which refers to variations in the proportion of hallucinated outputs generated in response to prompts in different languages.

Measuring these gaps can reveal that prompts in certain languages are more likely to induce hallucinations than others, which can significantly impact the reliability and trustworthiness of LLMs, especially in low-resource languages.

Previous work \citep{hong2024hallucinations, lin-etal-2022-truthfulqa} focused on measuring hallucinations through benchmarks that require human annotations, which can be costly and hard to scale for multilingual LLMs. These benchmarks are also not suited for a free-form text generation setup. As a result, an automated evaluation pipeline becomes highly desirable. 
Statistical measures like ROUGE fail to capture semantic variations \citep{sellam-etal-2020-bleurt} while NLI-based approaches transfer poorly to these tasks \citep{falke-etal-2019-ranking}. 
Among LLM-based methods, \citealp{chen2024inside} proposed an eigenvalue metric to measure self-consistent hallucination. However, measuring this metric is costly and might not be suitable for the evaluation of free-form generated text across multiple languages. 

Consequently, we explore \fs \citep{min-etal-2023-factscore}, a different LLM-based method to evaluate hallucinations. In particular, the \fs metric uses an LLM to fact-check outputs of other LLMs against a knowledge source. As the \fs was only developed and tested on English text, we extend the methodology to encompass different languages by comparing to knowledge sources in various languages and leveraging translation.

We evaluate LLMs from the LLaMA, Qwen and Aya families. We prompt them to generate biographies in 19 different languages and we then compute the \fs metric for each answer by comparing it to an external knowledge source, Wikipedia for this project. The computation is done through three different experimental setups. We finally analyzed the results with respect to the target language, to the experimental setup and to the LLM used for text generation. Our results show gaps in the \fs metric distribution across the prompt languages, particularly between high, medium, and low-resource languages. Our main contributions are: \begin{enumerate}[leftmargin=*]
\itemsep0em 
    \item Extending the \fs framework to a multilingual setting to quantify hallucination gaps across languages, with a focus on the disparities between high-resource and low-resource languages;
    \item Evaluating a range of open-source and multilingual models to investigate improvements associated with different architectures and model sizes;
    \item Assessing the robustness of the \fs framework across knowledge sources, prompt languages, and prompt templates.
\end{enumerate}

\section{Related work}

\paragraph{Evaluating hallucinations} Previous research has concentrated on evaluating, explaining, and mitigating hallucinations in language models \citep{ji2023survey, zhang2023sirens}.  All these efforts have been focused on detecting hallucinations in English-generated text.

There are several human-annotated benchmarks available for this purpose, including those compiled in the unified benchmark on HuggingFace by \citet{hong2024hallucinations}. 
Since these benchmarks rely on human annotation, they usually focus on short answers and are time-consuming to create, making them ill-suited for evaluating multilingual hallucinations in free-form text generation. 

Automatic metrics for measuring hallucinations encompass statistical and model-based ones \citep{ji2023survey}, many of which draw inspiration from summarization evaluation. Statistical metrics like ROUGE can only handle lexical information and fail to deal with syntactic or semantic variations \citep{sellam-etal-2020-bleurt}. NLI-based approaches are robust to lexical variability, but NLI models transfer poorly to abstractive summarization \citep{falke-etal-2019-ranking} and struggle to locate specific errors in generated content. Faithfulness Classification metrics \citep{liu-etal-2022-token} address this issue, but they rely heavily on English-annotated datasets. 

Among LLM-based methods, \citet{chen2024inside} proposed an eigenvalue-based metric for detecting self-consistent hallucinations. However, this approach is not well-suited for free-form text generation, where repeated prompts can produce different, yet correct, responses, leading to lower scores despite valid outputs. More relevant to long-form text generation are the methods proposed by \citet{min-etal-2023-factscore} and \citet{farquhar2024detecting}, both of which decompose answers into atomic facts. \citet{min-etal-2023-factscore} employs a LLM to fact-check these facts against a knowledge source, while \citet{farquhar2024detecting} uses semantic entropy probabilities. In this study, we adopt the \fs \citep{min-etal-2023-factscore} approach, computationally less expensive.

\paragraph{Multilingual LLM} Several studies have focused on evaluating language generation models within a multilingual framework. Some of these datasets include M3Exam \citep{zhang2023m3exam} for performance on human exam and Flores-101 \citep{goyal2022flores} for translation abilities. For the performance of the LLMs we used on these datasets, refer to Annex \ref{annex:llmsubj}. We can note that these evaluation metrics are still not consistently disclosed in technical reports or widely-recognized benchmarks. Despite covering a broad range of applications, these datasets do not cover hallucinations. However, they do provide evidence that LLMs exhibit different performance across different languages.

An additional open research question concerns how multilingual abilities in these models are acquired \citep{zhang-etal-2023-dont, wendler2024llamas}, as some models demonstrate proficiency in languages that are not officially supported. This observation motivated our decision to test models across a wide range of languages, even those not explicitly supported.

\paragraph{Hallucination metrics for multilingual generation} 
\citet{kang2024comparing} examine automatic hallucination detection metrics across different languages, including ROUGE, Named Entity Overlap, and the NLI-based SUMMAC score. Their findings show that these metrics do not correlate. Previous studies have suggested that these metrics may not be reliable for assessing hallucinations \citep{ji2023survey}, which motivates our investigation into LLM-based metrics, specifically the \fs metric, to evaluate hallucinations across languages on a range of open-source models.

The most related work to ours is \citet{shafayat2024multifact}, as the authors also study how to extend the \fs metric to a multilingual context. However, their methodology revolves around prompting in the original language and then translating generated content before assessing factuality.  We broaden our investigation by adding an experiment that prompts in English while requesting answers in another language, as well as another experiment that directly compares generations to the original language Wikipedia page. Further, we investigate the reliability of this choice of knowledge source. We also explore a wider range of languages, a different set of entities beyond politicians, as well as multilingual open-source models instead of ChatGPT. Additionally, our work critically examines the robustness of the metric itself and identifies areas for improving its reliability. 

\section{Measuring factuality}\label{factuality}

To evaluate factual hallucinations in multilingual free-form text generation, we use the \fs metric \citep{min-etal-2023-factscore}. This metric is particularly suited for our goal because it offers an intuitive, automated evaluation pipeline that can be easily adapted to different languages. By breaking down responses into atomic facts, \fs not only provides a more precise measure of factuality but also provides two key pieces of information: the factuality rate and the number of facts in the response.

Let's suppose we have a response $\mathcal{R}$ generated by an LLM, hereinafter referred to as \subjectLM. The \fs metric for this response $\mathcal{R}$ then consists of the following steps:
\begin{enumerate}[leftmargin=*]
\itemsep0em 
\item Decompose $\mathcal{R}$ into a set of atomic facts $\mathcal{A}(\mathcal{R})$. An atomic fact is a short sentence conveying a single piece of information. This is achieved by prompting an LLM, hereinafter referred to as \evalLM, to "Please breakdown the following sentence into independent facts" after showing it some decomposition examples. 
 \item Compare each fact $a \in \mathcal{A}(\mathcal{R})$ with an external knowledge source $\mathcal{C}$. To do so, we retrieve the proper passage from the source $\mathcal{C}$. We then construct a prompt by concatenating the retrieved passage, the given atomic fact and “True or False?”. We then feed this prompt to an LLM. We use the same LLM that was used to decompose atomic facts, the \evalLM. The answer gives us:
 $$\text{Supported}(a, \mathcal{C}) = \mathds{1}\{a \text{ supported by } \mathcal{C}\}$$ 
 \item (Optional) Add a length penalty $p$ depending on a hyperparameter $\gamma$ if our response $\mathcal{R}$ does not contain enough facts: 
 $$p=\exp\left(\frac{1-\gamma}{|\mathcal{A}|}\right) \quad \text{if } |\mathcal{A}(\mathcal{R})|\leq\gamma$$
For instance, without applying this penalty, a response containing only one correct fact would receive a 100\% \fs score, while a response with hundreds of facts, 99 of which are accurate, would get 99\%. We set the default parameter to $\gamma=10$, meaning responses with fewer than 10 facts are subject to a penalty.
 \item Compute the \fs $\text{F}(\mathcal{R}, \mathcal{C})$:
\begin{equation}\label{eq:fs}
    \text{F}(\mathcal{R}, \mathcal{C}) = \frac{p}{|\mathcal{A}(\mathcal{R})|} \sum_{a \in \mathcal{A}(\mathcal{R})}\text{Supported}(a, \mathcal{C}) 
\end{equation}
\end{enumerate}

\section{Methodology}
 In this section, we explain and discuss the experimental settings, i.e. the choices of \evalLM, \subjectLM, content to be generated and knowledge source, then detail the experimental process.
 
\subsection{Experimental settings}
Experiments are built following the methodology of the \fs paper \citep{min-etal-2023-factscore}. We prompt a \subjectLM to generate content in different languages and compute the \fs metric [\ref{eq:fs}] with an \evalLM for each answer, by comparing it to an external knowledge source, specifically Wikipedia for this project.

\paragraph{Choice of the \subjectLM}
Our objective is to evaluate open-source models with strong multilingual capabilities and developed in various countries. We also want to include for each model at least two different sizes to assess the impact of model size on the \fs. The final choice was set on LLaMA-3 (8B and 70B parameters), Aya-23 (8B and 35B parameters) and Qwen2 (7B and 72B parameters). Refer to Annex \ref{annex:llmsubj} for more details on the LLMs chosen. In the rest of the paper, we will refer to these models without specifying their versions.

\paragraph{Choice of the \evalLM}
To reduce bias in the evaluation process, we opt to use a different \evalLM than the \subjectLM models selected for the study. We use Mistral-7B-Instruct-v0.3 as the \evalLM and first compute the error rate and F1$_{micro}$ metrics (as detailed in \citealp{min-etal-2023-factscore}) by computing \fs with Mistral on human annotated data. 
In \citeauthor{min-etal-2023-factscore}, several methods are employed: \begin{itemize}[leftmargin=*]
\itemsep0em 
    \item No-context LM: Prompt "$<$atomic-fact$>$ True or False?";
    \item Retrieve$\rightarrow$LM: Retrieve passages from a knowledge source, concatenate these with the atomic fact and "True or False?" and prompt the concatenated result;
    \item Nonparametric Probability (NP): Mask each token in the atomic fact, calculate its likelihood with a nonparametric masked LM, average probabilities, and make a prediction based on thresholding;
    \item Retrieve$\rightarrow$LM + NP:  Assign "Supported" only if both Retrieve$\rightarrow$LM and NP assign "Supported".
\end{itemize} 

Based on their findings, we limit our experiments to the methods involving retrieval, comparing both Retrieve$\rightarrow$Mistral and Retrieve$\rightarrow$Mistral+NP. The results, presented in Table \ref{tab:llm-eval} in Annex \ref{annex:llmeval}, show that Mistral performs competitively compared to the models used in \citeauthor{min-etal-2023-factscore}, validating our choice of Mistral as \evalLM. However, unlike with the model Inst-LLaMA, adding NP did not enhance performance. We will thus use the Retrieve$\rightarrow$Mistral method for all subsequent experiments.

\paragraph{Prompts}

The \fs metric can be applied to any task, provided an appropriate knowledge source is available. We choose biographies because their factuality is easier to assess, as they generally include verifiable details such as birth dates and significant events, and they cover a wide range of nationalities. Besides, Wikipedia offers a multilingual knowledge source for this task, with biographies available in multiple languages. 

We now have to choose the people whose biographies we will ask for and in which languages. For language selection, our goal is to cover a range of languages that includes high-resource, medium-resource, and low-resource languages. We define languages categories, i.e. high, medium and low, based on their data ratios from the Common Crawl corpus,
drawing inspiration from \citealp{lai-etal-2023-chatgpt}. For example, Spanish and Chinese are high-resource languages, Persian and Hindi fall under medium-resource, and Tamil and Swahili are examples of low-resource languages. 

We also want to include as many language families as possible, while keeping the world's most widely spoken languages. Since we will be using Wikipedia as the knowledge source, it is also important to ensure that all selected languages have sufficient Wikipedia coverage, which we measure by the number of Wikipedia pages available in each language. Table \ref{tab:languages} in Annex \ref{annex:data} provides details on the 19 languages chosen, including the key statistics used to perform the selection. We do not take into account the languages supported by the \subjectLM in this selection, as some models demonstrate proficiency in languages that are not officially supported \citep{zhang-etal-2023-dont, wendler2024llamas}. We then proceed to select a set of notable figures with Wikipedia pages available in all these languages, resulting in a list of 485 individuals.
This selected set of entities is interestingly biased. Figures \ref{fig:citizen} and \ref{fig:lang} in Annex \ref{annex:data} illustrate their top 15 countries of citizenship and languages spoken, respectively. The distribution is largely skewed towards the American citizenship and English language.

\paragraph{Wikipedia as a knowledge source}
We retrieve Wikipedia summaries in every language for the 485 notable figures to serve as the knowledge source. Recognizing that Wikipedia content quality can vary across languages, we start by comparing these summaries to each other. For a given language, we translate the Wikipedia pages with GPT-4o-mini and we compute the \fs comparing to the English Wikipedia. We choose the English Wikipedia because the \fs indicates whether an atomic fact is supported by the knowledge source, rather than giving information on its presence in the knowledge source. Our assumption is that English Wikipedia is more comprehensive than other language versions. We also compute \fs for the English Wikipedia pages, i.e. comparing them against themselves. Results are presented in Figure \ref{fig:wikifs} in Annex \ref{annex:wiki}.

We can directly see that the evaluator is not perfect. Indeed, comparing the English Wikipedia to itself does not always yield a 100\% \fs, even if it typically falls within a high range of 90-100\%. For the other languages, cross-checking with English yields much lower \fs. This suggests that content in different languages can either contradict or diverge from what is found in English Wikipedia. These observations are important for the rest of our analysis. We will compare a generated biography with both its original language Wikipedia version and the English one. This may give us insight on how the LLM captures and represents knowledge in a multilingual setting, depending on whether the \fs is different when using these two different knowledge sources. It is also important to consider both for a more precise analysis. 

\subsection{Experiments}
In this section, we outline the experimental process, detailing the steps from data generation to \fs evaluation, with intermediate sanity checks. 

\paragraph{Data generation}
To ensure robustness in measuring the hallucination gaps, we use three established prompt templates for generating biographies from the literature: "Tell me a biography of \{\}", "Give me a biography of \{\}" and "Please give me a biography of \{\}". 

\noindent We use two prompting methods:
\begin{itemize}[leftmargin=*]
\itemsep0em 
    \item \texttt{lang}-prompt: Translate the template in the target language \texttt{lang} and use the translated prompt;
    \item \texttt{en}-prompt: Use the English template but add "in \{\texttt{lang}\}" at the end (e.g., "Tell me a biography of \{\} in French").
\end{itemize}
We use these methods for our six models \subjectLM, resulting in a total of 12 text generation setups. Each setup produces $485 \times 19 \times 3 $ generated responses. 

\paragraph{Sanity checks}
Once the biographies are generated, we carry out some sanity checks to ensure we have good enough generated answers to compute \fs with. 
For each answer, we verify that it is in the correct target language with the module \texttt{py3langid}. We set a threshold of 20 distinct words to remove outputs with same words repeated infinitely. We do not compute \fs for generated answers that do not pass sanity checks.

\paragraph{\fs evaluation}
We perform three experiments depending on the knowledge source and prompting method used. We will refer to these experiments as (prompt language, Wikipedia language):
\begin{enumerate}[leftmargin=*]
\itemsep0em 
    \item \multi: Compare the response produced with a \texttt{lang}-prompt to the \texttt{lang} Wikipedia page;
    \item \trans: Translate the response produced with a \texttt{lang}-prompt to English and compare to the English Wikipedia page;
    \item \enprompt: Translate the response produced with an \texttt{en}-prompt to English and compare to the English Wikipedia page.
\end{enumerate}

\noindent Recall that we generated three answers for each notable figure and language. We thus compute the \fs score for each answer and then average these three scores to obtain a single final score per entity. We also report the standard deviation. This process is repeated across all three experiments: \multi, \trans and \enprompt. 

\paragraph{Translation}
All translation steps, including prompt translations, generations, and Wikipedia pages, were performed with GPT-4.

\section{Results}

\subsection{Sanity checks}

The percentages of generated answers that passed sanity checks for each \subjectLM are shown in Table \ref{tab:sanitychecks} in Annex \ref{annex:sanity}. These initial results already demonstrate whether the \subjectLM is able to generate multilingual text, including in languages stated as not supported. 
Figure \ref{fig:langdetect} shows the percentage of generated responses produced in the correct target language, for each \subjectLM and prompt setting. As expected, the models generally perform better in generating text in high-resource languages compared to low- and very low-resource ones. For very low-resource languages like Javanese (jv) and Malay (ms), the models rarely generate text in the correct target language. 

The best performance overall is achieved with Qwen7 both in English (\texttt{en})  and original language (\texttt{lang}) prompting. We can note that for all models prompting in English tend to decrease the percentages of generated responses in the correct target language. Interestingly, for the LLaMA and Qwen families of models, increasing the number of parameters does not always lead to better performance, especially when prompted in English. 

We also observe poorer performance in the Japanese, Chinese, and Korean languages for the LLaMA models compared to others. When examining generated answers that failed the sanity checks, we notice that the LLaMA models often produced Romaji (writing in Roman characters) instead of Kanji (writing using Chinese characters). 

\subsection{\fs}
\noindent \textbf{The \subjectLM studied show different trends in hallucination gaps.} Figure \ref{fig:fs-enprompt} illustrates the mean \fs per model and per language for the \enprompt experiment. For results from the two other experiments, see Annex \ref{annex:exp-fs}. The model Qwen72 performs the best overall. This aligns with the models' multilingual performance on other benchmarks, where Qwen72 also ranks highest (see Table \ref{tab:multilingual}).

\begin{figure}[h!]
    \centering
    \begin{minipage}{0.45\textwidth}
        \centering
        \input{figure_scripts/lang_detect_heatmap}
        \vspace{-4mm}
        \caption{\centering Percentage of correct language produced per target language and generation set up}
        \label{fig:langdetect}
    \end{minipage}
    \hfill
    \begin{minipage}{0.45\textwidth}
        \centering
        \input{figure_scripts/fs_exp_enprompt}
        \vspace{-3mm}
        \caption{\centering \fs per language and per model for the \enprompt experiment}
        \label{fig:fs-enprompt}
    \end{minipage}
\end{figure}

Studying different subject models shows the impact of training sets, architectures, and model sizes on the \fs results. Aya and Qwen models, with extensive multilingual training data, have good performance in supported languages (see Annex \ref{annex:llmsubj}), outperforming LLaMA models, which are not optimized for multilingual tasks. Notably, Qwen models achieve the highest performance in Chinese, benefiting from tokenizers specifically designed for Chinese characters. Furthermore, within each model family, a higher number of parameters correlates with higher \fs. These findings reveal that multilingual hallucination gaps are influenced not only by high- versus low-resource language distinctions but also by factors like tokenizer design and language distribution in training data.

\begin{table*}[t]
    \center  
\resizebox{\linewidth}{!}{\begin{tabular}{lcccccc}
\toprule
\multirow{2}{*}{\textbf{Language Category}}& \multicolumn{3}{c}{\textbf{\fs (\%)}} & \multicolumn{3}{c}{\textbf{\# of Facts}} \\
\cmidrule(lr){2-4}  \cmidrule(lr){5-7}
 &  \enprompt & \trans & \multi & \enprompt & \trans & \multi \\
\midrule
Very-High &  73.7 \small{(± 10.1)} &   71.8 \small{(± 9.9)} &   70.3 \small{(± 9.7)} &            79 &    82 &   103 \\
High      &  70.2 \small{(± 12.6)} &  69.3 \small{(± 13.4)} &  58.5 \small{(± 16.4)} &            68 &    73 &    65 \\
Medium    &  64.7 \small{(± 16.0)} &  61.3 \small{(± 19.5)} &  47.8 \small{(± 19.4)} &            54 &    59 &    49 \\
Low       &  56.9 \small{(± 18.9)} &  47.6 \small{(± 23.4)} &  44.4 \small{(± 20.4)} &            38 &    34 &    53 \\

\bottomrule
\end{tabular}}
\caption{\centering Mean \fs (± STD) and Mean number of facts by Language Category and Experiment for all models}
\label{tab:fs-mean}
\end{table*}

\begin{figure*}[!h]
\centering
\input{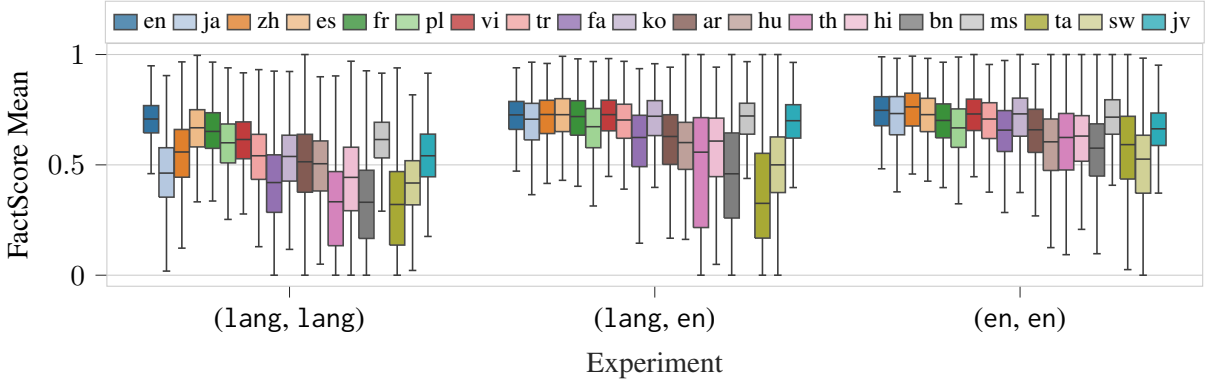}
\vspace{-6mm}
\caption{\centering FactScore Mean distribution by Language and Experiment for all models}
\label{fig:fs}
\end{figure*}

\noindent \textbf{Hallucination rates differ across languages.}  Table \ref{tab:fs-mean} presents the average results across all models and entities, grouped by language category. We observe multilingual hallucination gaps, with both factuality and the number of facts decreasing as the language resource level decline. Due to high standard deviations within these categories, we examine the \fs distribution at a finer level, for each language.

Figure \ref{fig:fs} shows these distributions across all models and entities, broken down by language and experiment. See Annex \ref{annex:all-fs} for the same results per \subjectLM. Aside from Malay (ms) and Javanese (jv) in all experiments, and Japanese (ja) and Chinese (zh) languages in the \multi experiment, we can observe the same trend as \fs distributions are more spread out and shifted toward lower values as the language resource level decline. It should be noted that after filtering out unsane answers that were in the incorrect language, we have less data points for the Malay and Javanese languages (see Figure \ref{fig:langdetect}). On the contrary, only for the \multi experiment, Japanese and Chinese show distributions that are more spread out and shifted toward lower values compared to other high-resource languages. This raises the question of why different experimental setups affect the results. \\

\noindent \textbf{Different pipelines show different results.} Table \ref{tab:fs-mean} and Figure \ref{fig:fs} highlight differences in results across the \multi, \trans, and \enprompt experiments, showing that the choice of knowledge source and prompt language can influence the \fs outcomes.

Regarding the prompt language, we gain insights into the performance of the \subjectLM. As shown in Figure \ref{fig:langdetect}, models respond differently to prompts given in English versus the target language \texttt{lang}. When comparing \trans and \enprompt — where the knowledge source remains the same but the prompt languages differ — we observe the highest results with the \enprompt setting. This raises concerns, as we would want models to respond accurately to prompts in original languages.

The impact of the knowledge source presents more significant challenges, as it directly affects the evaluator. In comparing \multi and \trans, where the prompt language remains  the same but the knowledge source differs, we see a decrease in \fs for the \multi experiment, and more dispersed distributions. This trend may be attributed to the quality of Wikipedia pages in their original languages, as highlighted in Figure \ref{fig:wikifs}. 

The performance of \multi compared to \trans and \enprompt is also influenced by the respective multilingual capabilities of the \evalLM and the translator (GPT-4). The \multi experiment is the only one that involves prompting the \evalLM in languages other than English. We noticed that for most languages, while breaking down a generation into atomic facts, the \evalLM also translated these facts into English even when not instructed to do so. Addressing this behavior might improve results in the \multi setting. The \trans approach seems to be the most suitable option, as it allows us to maintain prompts in their original languages. 
\\

\noindent \textbf{\fs's robustness depends on the language.}
Table \ref{tab:fs-std} in Annex \ref{annex:std-fs} presents \fs standard deviations across three prompt templates per entity. Averaging these by language category and experiment reveals that lower-resource languages exhibit higher \fs variability, indicating less consistency across prompt templates.\\ 

\section{Conclusion}
Our research shows multilingual hallucination gaps in LLMs: models tend to hallucinate more in low-resource languages and show greater factual accuracy in high-resource ones. Larger models generally perform better but still show hallucinations in low-resource languages. Even models with strong multilingual capabilities hallucinate in such languages, and struggle to generalize effectively to unsupported languages, suggesting that simply increasing model size or expanding training data is not a complete solution. This raises important concerns about the equitable performance of LLMs across different linguistic groups and the broader implications for fairness in AI technologies. 


\section*{Code availability}
The code and data are accessible at the GitHub repository: \url{https://github.com/cleachataig/Multilingual_Hallucination_Gaps}. 

\acks{Funding support for project activities has been partially provided by Canada CIFAR AI Chair, Facebook award, MEI award and FRQNT award. We also express our gratitude to Compute Canada and to Mila (mila.quebec) for their support in providing facilities for our evaluations.}

\bibliography{custom}

\newpage
\appendix

\section{Limitations}

The \fs metric, while useful for automatically assessing factuality in generated text, has several limitations that need to be addressed. One major issue is its robustness, especially in a multilingual setting. For instance, calculating a \fs using the generated text itself as the knowledge source does not always yield a perfect 100\% score, as shown on Figure \ref{fig:wikifs}, highlighting potential inconsistencies. 
Computing \fs can be quite resource-intensive, which can limit its widespread use.
Besides, we can only verify \textit{intrinsic} hallucinations  \citep{ji2023survey} with this metric, i.e. when the generated content directly contradicts the reference. Future work could extend the metric to \textit{extrinsic} hallucinations, when the generated output cannot be verified with the source reference (i.e., it is neither supported nor contradicted by the reference), to provide more insights. This would be useful for languages in which the Wikipedia coverage may be weaker than the English one. 

The use of Wikipedia as a knowledge source also presents limitations. Some Wikipedia entries may not be fully accurate, and certain facts could be ambiguous. Wikipedia's coverage also varies significantly across languages, which can impact the effectiveness of the \fs metric in the \multi setting. Despite these issues, Wikipedia remains one of the most comprehensive public multilingual knowledge sources available.  

Another limitation comes from potential biases in the \fs metric evaluation, as the computation is done by another LLM, the \evalLM. This is especially evident in a multilingual setup, as for the \multi experiment the evaluation relies on the performance of the \evalLM across languages. We assume equal performance across languages, which is not accurate in practice. For the other experiments, we rely on GPT4's performance in translation tasks which can also add variability. To address this, creating a human benchmark for assessing multilingual hallucination gaps could offer a more reliable and unbiased evaluation.

Finally, we only focus on biographies of a specific group of individuals. While we cover a diverse set of people, future work could explore how these gaps evolve when the \subjectLM are confronted with other tasks, for instance other types of articles on Wikipedia (e.g., scientific topics) or text about historical events whose knowledge source can be a collection of articles. However, for consistency with our experimental settings, these tasks would need to have multilingual knowledge sources for evaluation and less room for subjectivity.

\section{Characteristics of the \subjectLM} \label{annex:llmsubj}
Table \ref{tab:multilingual}  presents the models chosen as \subjectLM, as well as their characteristics and their multilingual performance on different benchmarks, as reported in LLaMA-3 \citep{dubey2024llama3herdmodels}, Qwen2 \citep{qwen2} and Aya-23 \citep{aryabumi2024aya23openweight} technical reports.
\begin{table}[!ht]
\centering
\begin{tabular}{lcccccc}
\toprule
 &  \multicolumn{2}{c}{Aya-23} & \multicolumn{2}{c}{LLaMA-3} & \multicolumn{2}{c}{Qwen2}\\
\cmidrule(lr){2-3} \cmidrule(lr){4-5} \cmidrule(lr){6-7}
\# Parameters & 8B & 35B & 8B & 70B & 7B & 72B \\
\midrule
Architecture & \multicolumn{2}{c}{Dense} & \multicolumn{2}{c}{Dense} & \multicolumn{2}{c}{Dense} \\
Developer & \multicolumn{2}{c}{Cohere} & \multicolumn{2}{c}{Meta-AI} & \multicolumn{2}{c}{Alibaba} \\
Origin & \multicolumn{2}{c}{Canada} & \multicolumn{2}{c}{USA} & \multicolumn{2}{c}{China} \\
\midrule
Exam\textsuperscript{1} &  48 & 58 & 52 & 70 & 60 & \textbf{78} \\
Understanding\textsuperscript{2} & - &- &69&80&72&\textbf{81} \\
Mathematics\textsuperscript{3} & 37 & 47 & 36 & 67 & 57 & \textbf{87} \\
Translation\textsuperscript{4} & 37 & \textbf{40} & 32 & 38 & 32 & 38 \\
\bottomrule

\multicolumn{7}{l}{\tiny{\textsuperscript{1} mMMLU \, 
\textsuperscript{2} BELEBELE, XCOPA, XWinograd, XStoryCloze, PAWS-X \,  \textsuperscript{3} MGSM \, 
\textsuperscript{4} Flores-101}}
\end{tabular}
\vspace{2mm}
\caption{Characteristics and Multilingual Performance of the Large Language Models chosen as \subjectLM. The \textbf{bold} values indicate the best performance for each multilingual benchmark. }
\label{tab:multilingual}
\end{table}

LlaMA officially supports 8 languages \citep{dubey2024llama3herdmodels}: English, German, French, Italian, Portuguese, Hindi, Spanish, and Thai, although the underlying foundation model has been trained on a broader collection of languages. 

The Aya models \citep{aryabumi2024aya23openweight} support only a limited set of languages that intersect with ours: English (en), Japanese (ja), Chinese (zh), Spanish (es), French (fr), Polish (pl), Vietnamese (vi), Turkish (tr), Persian (fa), Korean (ko), Arabic (ar), and Hindi (hi). 

The Qwen models \citep{qwen2} are the ones covering the most languages of our dataset, with the exception of Hungarian (hu), Tamil (ta), Swahili (sw) and Javanese (jv).

\section{Validation of the \evalLM}\label{annex:llmeval} 
Table \ref{tab:llm-eval} presents validation results for the \fs estimated by Mistral compared to human annotated scores. We also include results of the two best models of \citeauthor{min-etal-2023-factscore}.
\begin{table*}[!ht]
    \center  
    \begin{tabular}{lcccccc}
        \toprule
            \multirow{2}{*}{\textbf{\evalLM}} & \multicolumn{2}{c}{\textsc{subj}: InstGPT} & \multicolumn{2}{c}{\textsc{subj}: ChatGPT} & \multicolumn{2}{c}{\textsc{subj}: PPLAI} \\
            \cmidrule(lr){2-3} \cmidrule(lr){4-5} \cmidrule(lr){6-7}
            & ER & F1 & ER & F1 & ER & F1 \\
            \midrule
            Always Not-supported & 0.42 & 71.4 & 0.58 & 58.3 & 0.80 & 30.9 \\
            \midrule
            Retrieve$\rightarrow$ChatGPT    & 0.14 & \textbf{86.2} & 0.18 & 68.5 & \textbf{0.09} & 54.9 \\
            Retrieve$\rightarrow$Inst-LLaMA+NP    & 0.22 & 73.3 & 0.29 & 60.2 & 0.36 & 39.6 \\
            \midrule
            Retrieve$\rightarrow$Mistral    & \textbf{0.09} & 85.4 & \textbf{0.11} & 73.5 &  0.11 & \textbf{58.4} \\
            Retrieve$\rightarrow$Mistral+NP  & 0.11 & 84.8 & 0.12 & \textbf{74.0} & 0.17 & 56.3 \\
        \bottomrule
    \end{tabular}
    \caption{
        Results on Error Rate (ER) and F1$_{micro}$ (F1) for the \fs estimated by Mistral compared to human annotated scores. We also include results of the 2 best models of \citeauthor{min-etal-2023-factscore}. The \textbf{bold} values indicate the best performance for each metric.
    }\label{tab:llm-eval}
\end{table*}

\section{Languages and People Dataset}\label{annex:data}
Table \ref{tab:languages} present the 19 selected languages along with their characteristics used for the selection process.

Figures \ref{fig:citizen} and \ref{fig:lang} illustrate the top 15 countries of citizenship and languages of the entities. It is important to note that a figure may have multiple citizenships or speak multiple languages. We can observe that the data distribution is largely skewed towards the American citizenship and the English language.

 \begin{table*}[htbp]
    \centering
    \resizebox{\linewidth}{!}{
            \begin{tabular}{cccccc}
      \toprule
       & Family & Branch & CC Ratio & \makecell{Worldwide Speakers \\(in millions)} & \makecell{Wikipedia pages \\(in thousands)}\\
      \midrule
       \multicolumn{6}{c}{Very High Resource Language} \\ 
      English (en)  & Indo-European & Germanic & 46.45 & 1,456  &  6,832     \\ \midrule
      \multicolumn{6}{c}{High Resource Languages} \\ 
      Japanese (ja) &Japonic & - &5.09	&123 &1,419\\
      Chinese (zh)    & Sino-Tibetan & Sinitic & 4.17    & 1,138       &  1,423   \\
      Spanish (es)  & Indo-European &Romance & 4.55	           & 559       & 1,957      \\
    French (fr) & Indo-European & Romance & 4.64      & 310 & 2,616      \\
    Polish (pl) &Indo-European&Balto-Slavic&1.76&41&1,620\\
   \midrule
   \multicolumn{6}{c}{Medium Resource Languages} \\ 
    Vietnamese (vi) & Austroasiatic & Vietic & 0.99&86&  1,294\\
      Turkish (tr) & Turkic & Oghuz & 0.99	         & 90     & 608      \\
      Persian (fa) &Indo-European&Iranian& 0.67 &79& 1,004\\
      Korean (ko) &Koreanic&-& 0.65&82&672\\
    Arabic (ar) & Afro-Asiatic & Semitic & 0.59         & 274    & 1,235      \\
    Hungarian (hu) & Uralic & Hungarian & 0.56 & 17 & 543\\
    Thai (th) & Kra–Dai & Zhuang–Tai & 0.41 & 61 & 165\\
      Hindi (hi)  & Indo-European & Indo-Aryan & 0.18          & 610     & 162     \\\midrule
       \multicolumn{6}{c}{Low and Very-Low Resource Languages} \\ 
      Bengali (bn) & Indo-European & Indo-Aryan & 0.10 & 273 &  154\\
      Malay (ms)  & Austronesian & Malay & 0.07      & 290     & 377     \\
      Tamil (ta) &Dravidian& Southern & 0.04 & 87  & 166\\
      Swahili (sw)  & Niger-Congo & Bantu &  0.008       & 72     & 80    \\
       Javanese (jv) & Austronesian & Malayo-Polynesian & 0.002      & 68     & 73 \\
      \bottomrule
    \end{tabular}}
    \caption{The 19 chosen languages with key statistics}
    \label{tab:languages}
  \end{table*}

\begin{figure}[h!]
    \centering
    \begin{minipage}{0.45\textwidth}
        \centering
        \input{figure_scripts/citizenships}
        \caption{Top 15 citizenship countries}
        \label{fig:citizen}
    \end{minipage}
    \hfill
    \begin{minipage}{0.45\textwidth}
        \centering
        \input{figure_scripts/languages}
        \caption{Top 15 languages spoken}
        \label{fig:lang}
    \end{minipage}
    \caption{Citizenship and language statistics of the entities}
    \label{fig:citizenship_languages}
\end{figure}

\section{Wikipedia \fs} \label{annex:wiki}
Figure \ref{fig:wikifs} present the \fs distribution results when comparing the different Wikipedia summaries in every language to the English Wikipedia one. 

We can see that the evaluator is not perfect, since comparing the English Wikipedia to itself does not always yield a 100\% \fs, even if it typically falls within a high range of 90-100\%. For the other languages, cross-checking with English yields much lower \fs. This suggest that content in different languages can either contradict or diverge from what is found in English Wikipedia. 

\begin{figure}[!ht]
\centering
\input{figure_scripts/wiki_factscores}
\caption{\fs distribution for Wikipedia pages}
\label{fig:wikifs}
\end{figure}

\section{Sanity checks}\label{annex:sanity}

 Table \ref{tab:sanitychecks} show the percentages of generated answers that passed sanity checks for each \subjectLM. Overall, the percentages are similar across across all \subjectLM.

For the LLaMA models, we initially encountered lower percentages of sane answers (22\%) for the \texttt{lang} prompt. When prompted in another language than English, these models failed to understand that they had to respond in that language and not in English. To address this issue, we translated the entire English prompt, with the added directive "in \{\texttt{lang}\}". For example, for the French language, the prompt was adjusted to "Donne-moi une biographie de \{\} en Français." instead of only "Donne-moi une biographie de \{\}.". This adjustment significantly improved the percentage of sane answers, raising it from 22\% to 88\%. No additional output regeneration was needed for the other models, as they already produced satisfactory percentages of sane responses.

  \begin{table}[H]
    \center  
    \begin{tabular}{lcc}
        \toprule
            &\texttt{en}-prompt & \texttt{lang}-prompt \\
             \midrule
            LLaMA-3 8B &83.68 & 88.50\\
            LLaMA-3 70B   & 81.55  & 94.96\\\midrule
            Qwen 7B    & 89.76 & 89.08\\
            Qwen 72B    & 74.48  & 89.44\\\midrule
            Aya 8B    & 70.79 & 80.42\\
            Aya 35B   &84.52  & 83.85 \\
        \bottomrule
    \end{tabular}
    \vspace{3mm}
    \caption{
        Percentages of generated answers kept after sanity checks per generation setup
    }\label{tab:sanitychecks}
\end{table}

\section{\fs results per experiment}\label{annex:exp-fs}
Figures \ref{fig:fs-multi} and \ref{fig:fs-trans} present the mean \fs per model and per language for the \multi and \trans experiments. We observe the same trends across \subjectLM as for the \enprompt experiment.

\begin{figure}[h!]
    \centering
    \begin{minipage}{0.45\textwidth}
        \centering
        \input{figure_scripts/fs_exp_multi}
        \vspace{-3mm}
        \caption{\multi experiment}
        \label{fig:fs-multi}
    \end{minipage}
    \hfill
    \begin{minipage}{0.45\textwidth}
        \centering
        \input{figure_scripts/fs_exp_trans}
        \vspace{-3mm}
        \caption{\trans experiment}
        \label{fig:fs-trans}
    \end{minipage}
\end{figure}

\section{Standard deviation of \fs}\label{annex:std-fs}
Table \ref{tab:fs-std} presents \fs standard deviations across three prompt templates per entity.
\begin{table}[!h]
    \center  
\begin{tabular}{lccc}
\toprule
\multirow{2}{*}{\textbf{Language Category}}& \multicolumn{3}{c}{\textbf{\fs (\%)}} \\
\cmidrule(lr){2-4} 
&  \enprompt & \trans & \multi\\
\midrule
Very-High &      4.9 &   5.1 &   4.8  \\
High      &      6.2 &   6.6 &   7.0 \\
Medium    &      7.5 &   8.0 &   8.6 \\
Low       &      8.8 &  10.2 &   9.3 \\
\bottomrule
\end{tabular}
\caption{\centering Standard deviation across the 3 prompt templates of \fs by Language Category and Experiment for all models}
\label{tab:fs-std}
\end{table}

  \section{\fs results per \subjectLM}\label{annex:all-fs}
We present in this section the \fs results for every \subjectLM instead of the average over all models. For every \subjectLM we present both the table of \fs and number of facts averaged across language categories and the boxplot figures of distribution for each language.

  \begin{table*}[!ht]
    \center  
\resizebox{\linewidth}{!}{\begin{tabular}{lcccccc}
\toprule
\multirow{2}{*}{\textbf{Language Category}}& \multicolumn{3}{c}{\textbf{\fs (\%)}} & \multicolumn{3}{c}{\textbf{\# of Facts}} \\
\cmidrule(lr){2-4}  \cmidrule(lr){5-7}
 &  \enprompt & \trans & \multi & \enprompt & \trans & \multi \\
\midrule
Very-High &  72.6 \small{(± 12.8)} &  70.2 \small{(± 12.5)} &  70.1 \small{(± 12.6)} &            67 &    75 &    82 \\
High      &  70.6 \small{(± 13.9)} &  68.3 \small{(± 15.4)} &  60.8 \small{(± 19.4)} &            68 &    80 &    54 \\
Medium    &  59.6 \small{(± 20.7)} &  55.0 \small{(± 23.4)} &  43.7 \small{(± 23.9)} &            51 &    58 &    36 \\
Low       &  32.9 \small{(± 23.7)} &  28.4 \small{(± 17.3)} &  41.7 \small{(± 26.0)} &            26 &    19 &    36 \\
\bottomrule
\end{tabular}}
\vspace{3mm}
\caption{\centering Mean \fs (± STD) and Mean number of facts by Language Category and Experiment for Aya 8}
\label{tab:fs-aya8}
\end{table*}

  \begin{figure*}[!ht]
\centering
\input{figure_scripts/fs_aya_8}
\vspace{-6mm}
\caption{\centering FactScore Mean distribution by Language and Experiment for Aya 8}
\label{fig:fs-aya8}
\end{figure*}

\begin{table*}[!ht]
    \center  
\resizebox{\linewidth}{!}{\begin{tabular}{lcccccc}
\toprule
\multirow{2}{*}{\textbf{Language Category}}& \multicolumn{3}{c}{\textbf{\fs (\%)}} & \multicolumn{3}{c}{\textbf{\# of Facts}} \\
\cmidrule(lr){2-4}  \cmidrule(lr){5-7}
 &  \enprompt & \trans & \multi & \enprompt & \trans & \multi \\
\midrule
Very-High &   76.4 \small{(± 9.8)} &   75.1 \small{(± 9.6)} &   74.7 \small{(± 9.6)} &            83 &    80 &    89 \\
High      &  75.4 \small{(± 11.1)} &  74.4 \small{(± 12.6)} &  65.0 \small{(± 17.7)} &            78 &    80 &    59 \\
Medium    &  67.0 \small{(± 17.1)} &  63.2 \small{(± 23.4)} &  49.5 \small{(± 23.2)} &            58 &    60 &    40 \\
Low       &  52.2 \small{(± 22.6)} &  28.3 \small{(± 21.7)} &  38.0 \small{(± 26.2)} &            26 &    21 &    37 \\
\bottomrule
\end{tabular}}
\vspace{3mm}
\caption{\centering Mean \fs (± STD) and Mean number of facts by Language Category and Experiment for Aya 35}
\label{tab:fs-aya35}
\end{table*}

  \begin{figure*}[!ht]
\centering
\input{figure_scripts/fs_aya_35}
\vspace{-6mm}
\caption{\centering FactScore Mean distribution by Language and Experiment for Aya 35}
\label{fig:fs-aya35}
\end{figure*}

\begin{table*}[!ht]
    \center  
\resizebox{\linewidth}{!}{\begin{tabular}{lcccccc}
\toprule
\multirow{2}{*}{\textbf{Language Category}}& \multicolumn{3}{c}{\textbf{\fs (\%)}} & \multicolumn{3}{c}{\textbf{\# of Facts}} \\
\cmidrule(lr){2-4}  \cmidrule(lr){5-7}
 &  \enprompt & \trans & \multi & \enprompt & \trans & \multi \\
\midrule
Very-High &   71.2 \small{(± 9.0)} &   70.5 \small{(± 8.5)} &   69.9 \small{(± 8.1)} &            76 &    82 &   107 \\
High      &  61.8 \small{(± 11.9)} &  63.7 \small{(± 13.8)} &  56.7 \small{(± 13.7)} &            60 &    66 &    73 \\
Medium    &  59.8 \small{(± 11.6)} &  58.4 \small{(± 16.2)} &  48.1 \small{(± 16.5)} &            55 &    65 &    57 \\
Low       &  59.7 \small{(± 12.8)} &  54.3 \small{(± 18.8)} &  46.0 \small{(± 16.6)} &            39 &    43 &    56 \\
\bottomrule
\end{tabular}}
\vspace{3mm}
\caption{\centering Mean \fs (± STD) and Mean number of facts by Language Category and Experiment for LLaMA 8}
\label{tab:fs-llama8}
\end{table*}

  \begin{figure*}[!ht]
\centering
\input{figure_scripts/fs_llama_8}
\vspace{-6mm}
\caption{\centering FactScore Mean distribution by Language and Experiment for LLaMA 8}
\label{fig:fs-llama8}
\end{figure*}

\begin{table*}[!ht]
    \center  
\resizebox{\linewidth}{!}{\begin{tabular}{lcccccc}
\toprule
\multirow{2}{*}{\textbf{Language Category}}& \multicolumn{3}{c}{\textbf{\fs (\%)}} & \multicolumn{3}{c}{\textbf{\# of Facts}} \\
\cmidrule(lr){2-4}  \cmidrule(lr){5-7}
 &  \enprompt & \trans & \multi & \enprompt & \trans & \multi \\
\midrule
Very-High &   70.7 \small{(± 8.7)} &   66.7 \small{(± 8.1)} &   67.1 \small{(± 7.9)} &            81 &    92 &   120 \\
High      &  67.1 \small{(± 10.7)} &  68.6 \small{(± 11.5)} &  56.0 \small{(± 13.4)} &            66 &    69 &    74 \\
Medium    &   68.2 \small{(± 9.8)} &  68.1 \small{(± 11.5)} &  51.5 \small{(± 15.4)} &            64 &    68 &    63 \\
Low       &  67.3 \small{(± 11.1)} &  67.4 \small{(± 15.6)} &  49.8 \small{(± 14.4)} &            45 &    48 &    66 \\
\bottomrule
\end{tabular}}
\vspace{3mm}
\caption{\centering Mean \fs (± STD) and Mean number of facts by Language Category and Experiment for LLaMA 70}
\label{tab:fs-llama70}
\end{table*}

  \begin{figure*}[!ht]
\centering
\input{figure_scripts/fs_llama_70}
\vspace{-6mm}
\caption{\centering FactScore Mean distribution by Language and Experiment for LLaMA 70}
\label{fig:fs-llama70}
\end{figure*}

\begin{table*}[!ht]
    \center  
\resizebox{\linewidth}{!}{\begin{tabular}{lcccccc}
\toprule
\multirow{2}{*}{\textbf{Language Category}}& \multicolumn{3}{c}{\textbf{\fs (\%)}} & \multicolumn{3}{c}{\textbf{\# of Facts}} \\
\cmidrule(lr){2-4}  \cmidrule(lr){5-7}
 &  \enprompt & \trans & \multi & \enprompt & \trans & \multi \\
\midrule
Very-High &   74.7 \small{(± 9.2)} &   73.0 \small{(± 9.2)} &   70.1 \small{(± 9.4)} &            81 &    83 &   108 \\
High      &  68.3 \small{(± 11.7)} &  65.6 \small{(± 12.2)} &  53.3 \small{(± 15.0)} &            66 &    70 &    65 \\
Medium    &  59.3 \small{(± 15.2)} &  52.8 \small{(± 17.5)} &  41.9 \small{(± 17.5)} &            50 &    51 &    48 \\
Low       &  43.3 \small{(± 15.3)} &  37.8 \small{(± 17.3)} &  42.2 \small{(± 16.7)} &            36 &    32 &    62 \\
\bottomrule
\end{tabular}}
\vspace{3mm}
\caption{\centering Mean \fs (± STD) and Mean number of facts by Language Category and Experiment for Qwen 7}
\label{tab:fs-qwen7}
\end{table*}

  \begin{figure*}[!ht]
\centering
\input{figure_scripts/fs_qwen_7}
\vspace{-6mm}
\caption{\centering FactScore Mean distribution by Language and Experiment for Qwen 7}
\label{fig:fs-qwen7}
\end{figure*}

\begin{table*}[!ht]
    \center  
\resizebox{\linewidth}{!}{\begin{tabular}{lcccccc}
\toprule
\multirow{2}{*}{\textbf{Language Category}}& \multicolumn{3}{c}{\textbf{\fs (\%)}} & \multicolumn{3}{c}{\textbf{\# of Facts}} \\
\cmidrule(lr){2-4}  \cmidrule(lr){5-7}
 &  \enprompt & \trans & \multi & \enprompt & \trans & \multi \\
\midrule
Very-High &   76.6 \small{(± 8.8)} &   73.5 \small{(± 8.3)} &   70.2 \small{(± 8.5)} &            84 &    86 &   109 \\
High      &   75.8 \small{(± 9.9)} &   74.7 \small{(± 9.6)} &  59.3 \small{(± 15.5)} &            66 &    71 &    62 \\
Medium    &  74.4 \small{(± 11.7)} &  72.3 \small{(± 11.3)} &  52.2 \small{(± 15.8)} &            48 &    52 &    49 \\
Low       &  67.5 \small{(± 13.8)} &  57.4 \small{(± 19.7)} &  48.7 \small{(± 15.8)} &            43 &    29 &    62 \\
\bottomrule
\end{tabular}}
\vspace{3mm}
\caption{\centering Mean \fs (± STD) and Mean number of facts by Language Category and Experiment for Qwen 72}
\label{tab:fs-qwen72}
\end{table*}

  \begin{figure*}[!ht]
\centering
\input{figure_scripts/fs_qwen_72}
\vspace{-6mm}
\caption{\centering FactScore Mean distribution by Language and Experiment for Qwen 72}
\label{fig:fs-qwen72}
\end{figure*}
\end{document}

%% file: figure_scripts/lang_detect_heatmap.tex
\begin{tikzpicture}

\definecolor{darkgray176}{RGB}{176,176,176}

\begin{axis}[width=\linewidth, height=\linewidth,
colorbar,
colormap={mymap}{[1pt]
  rgb(0pt)=(0.705673158,0.01555616,0.150232812);
  rgb(1pt)=(0.752534934,0.157246067,0.184115123);
  rgb(2pt)=(0.795631745,0.24128379,0.220525627);
  rgb(3pt)=(0.834620542,0.312874446,0.259301199);
  rgb(4pt)=(0.869186849,0.378313092,0.300267182);
  rgb(5pt)=(0.89904617,0.439559467,0.343229596);
  rgb(6pt)=(0.923944917,0.49730856,0.387970225);
  rgb(7pt)=(0.943660866,0.551750968,0.434243684);
  rgb(8pt)=(0.958003065,0.602842431,0.481775914);
  rgb(9pt)=(0.966811177,0.650421156,0.530263762);
  rgb(10pt)=(0.969954137,0.694266682,0.579375448);
  rgb(11pt)=(0.96732803,0.734132809,0.628751763);
  rgb(12pt)=(0.958852946,0.769767752,0.678007945);
  rgb(13pt)=(0.944468518,0.800927443,0.726736146);
  rgb(14pt)=(0.924127593,0.827384882,0.774508472);
  rgb(15pt)=(0.897787179,0.848937047,0.820880546);
  rgb(16pt)=(0.865395197,0.86541021,0.865395561);
  rgb(17pt)=(0.833466556,0.860207984,0.901068838);
  rgb(18pt)=(0.798691636,0.849786142,0.931688648);
  rgb(19pt)=(0.761464949,0.834302879,0.956945269);
  rgb(20pt)=(0.722193294,0.813952739,0.976574709);
  rgb(21pt)=(0.681291281,0.788964712,0.990363227);
  rgb(22pt)=(0.639176211,0.759599947,0.998151185);
  rgb(23pt)=(0.596262162,0.726149107,0.999836203);
  rgb(24pt)=(0.552953156,0.688929332,0.995375608);
  rgb(25pt)=(0.509635204,0.648280772,0.98478814);
  rgb(26pt)=(0.46666708,0.604562568,0.968154911);
  rgb(27pt)=(0.424369608,0.558148092,0.945619588);
  rgb(28pt)=(0.38301334,0.50941904,0.917387822);
  rgb(29pt)=(0.342804478,0.458757618,0.883725899);
  rgb(30pt)=(0.30386891,0.406535296,0.84495867);
  rgb(31pt)=(0.26623388,0.353094838,0.801466763);
  rgb(32pt)=(0.2298057,0.298717966,0.753683153)
},
colorbar/width=2.5mm,
point meta max=100,
point meta min=0,
tick align=outside,
tick pos=left,
x grid style={darkgray176},
xlabel={Generation set up},
xlabel style={yshift=-6ex},
xmin=0, xmax=12,
xtick style={color=black},
xtick={0.5,1.5,2.5,3.5,4.5,5.5,6.5,7.5,8.5,9.5,10.5,11.5},
    xticklabels={
      \parbox{0.2cm}{\rotatebox{90}{Lang} \\ Aya8}, 
      \parbox{0.2cm}{\rotatebox{90}{En}\\}, 
      \parbox{0.2cm}{\rotatebox{90}{Lang} \\ Aya35}, 
      \parbox{0.2cm}{\rotatebox{90}{En} \\ }, 
      \parbox{0.2cm}{\rotatebox{90}{Lang} \\ LLaMA\\8}, 
      \parbox{0.2cm}{\rotatebox{90}{En} \\ }, 
      \parbox{0.2cm}{\rotatebox{90}{Lang} \\ LLaMA \\ 70}, 
      \parbox{0.2cm}{\rotatebox{90}{En} \\ }, 
      \parbox{0.2cm}{\rotatebox{90}{Lang} \\ Qwen7}, 
      \parbox{0.2cm}{\rotatebox{90}{En} \\ }, 
      \parbox{0.2cm}{\rotatebox{90}{Lang} \\ Qwen72}, 
      \parbox{0.2cm}{\rotatebox{90}{En} \\ }, 
    },
    xticklabel style={align=center},
y dir=reverse,
y grid style={darkgray176},
ylabel={Target Language},
ylabel near ticks,
ylabel style={xshift=-1ex},
ymin=0, ymax=19,
ytick style={color=black},
ytick={0.5,1.5,2.5,3.5,4.5,5.5,6.5,7.5,8.5,9.5,10.5,11.5,12.5,13.5,14.5,15.5,16.5,17.5,18.5},
yticklabels={en, ja, zh, es, fr, pl, vi, tr, fa, ko, ar, hu, th, hi, bn, ms, ta, sw, jv},
ticklabel style={font=\tiny},
]
\addplot graphics [includegraphics cmd=\pgfimage,xmin=0, xmax=12, ymin=19, ymax=0] {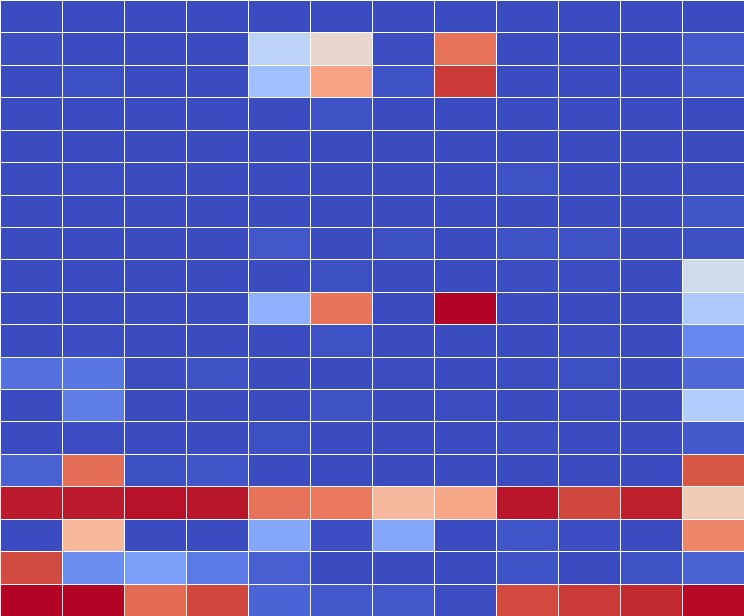};

\end{axis}

\end{tikzpicture}

%% file: figure_scripts/fs_exp_enprompt.tex
\begin{tikzpicture}

\definecolor{darkslategray38}{RGB}{38,38,38}
\definecolor{lightgray204}{RGB}{204,204,204}

\begin{axis}[width=\linewidth, height=\linewidth,
axis line style={lightgray204},
colorbar,
colormap={mymap}{[1pt]
  rgb(0pt)=(0.705673158,0.01555616,0.150232812);
  rgb(1pt)=(0.752534934,0.157246067,0.184115123);
  rgb(2pt)=(0.795631745,0.24128379,0.220525627);
  rgb(3pt)=(0.834620542,0.312874446,0.259301199);
  rgb(4pt)=(0.869186849,0.378313092,0.300267182);
  rgb(5pt)=(0.89904617,0.439559467,0.343229596);
  rgb(6pt)=(0.923944917,0.49730856,0.387970225);
  rgb(7pt)=(0.943660866,0.551750968,0.434243684);
  rgb(8pt)=(0.958003065,0.602842431,0.481775914);
  rgb(9pt)=(0.966811177,0.650421156,0.530263762);
  rgb(10pt)=(0.969954137,0.694266682,0.579375448);
  rgb(11pt)=(0.96732803,0.734132809,0.628751763);
  rgb(12pt)=(0.958852946,0.769767752,0.678007945);
  rgb(13pt)=(0.944468518,0.800927443,0.726736146);
  rgb(14pt)=(0.924127593,0.827384882,0.774508472);
  rgb(15pt)=(0.897787179,0.848937047,0.820880546);
  rgb(16pt)=(0.865395197,0.86541021,0.865395561);
  rgb(17pt)=(0.833466556,0.860207984,0.901068838);
  rgb(18pt)=(0.798691636,0.849786142,0.931688648);
  rgb(19pt)=(0.761464949,0.834302879,0.956945269);
  rgb(20pt)=(0.722193294,0.813952739,0.976574709);
  rgb(21pt)=(0.681291281,0.788964712,0.990363227);
  rgb(22pt)=(0.639176211,0.759599947,0.998151185);
  rgb(23pt)=(0.596262162,0.726149107,0.999836203);
  rgb(24pt)=(0.552953156,0.688929332,0.995375608);
  rgb(25pt)=(0.509635204,0.648280772,0.98478814);
  rgb(26pt)=(0.46666708,0.604562568,0.968154911);
  rgb(27pt)=(0.424369608,0.558148092,0.945619588);
  rgb(28pt)=(0.38301334,0.50941904,0.917387822);
  rgb(29pt)=(0.342804478,0.458757618,0.883725899);
  rgb(30pt)=(0.30386891,0.406535296,0.84495867);
  rgb(31pt)=(0.26623388,0.353094838,0.801466763);
  rgb(32pt)=(0.2298057,0.298717966,0.753683153)
},
colorbar/width=2.5mm,
point meta max=0.800892984278599,
point meta min=0.272551951555632,
tick align=outside,
tick pos=left,
x grid style={lightgray204},
xlabel={Model},
xlabel style={yshift=-7ex},
xmin=0, xmax=6,
xtick style={color=darkslategray38},
xtick={0.5,1.5,2.5,3.5,4.5,5.5},
xticklabels={Aya8, Aya35, LLaMA8, LLaMA70, Qwen7, Qwen72},
xticklabel style={rotate=45.0},
y dir=reverse,
y grid style={lightgray204},
ylabel={Language},
ymin=0, ymax=19,
ytick style={color=darkslategray38},
ytick={0.5,1.5,2.5,3.5,4.5,5.5,6.5,7.5,8.5,9.5,10.5,11.5,12.5,13.5,14.5,15.5,16.5,17.5,18.5},
yticklabels={en, ja, zh, es, fr, pl, vi, tr, fa, ko, ar, hu, th, hi, bn, ms, ta, sw, jv},
ticklabel style={font=\scriptsize},
]
\addplot graphics [includegraphics cmd=\pgfimage,xmin=0, xmax=6, ymin=19, ymax=0] {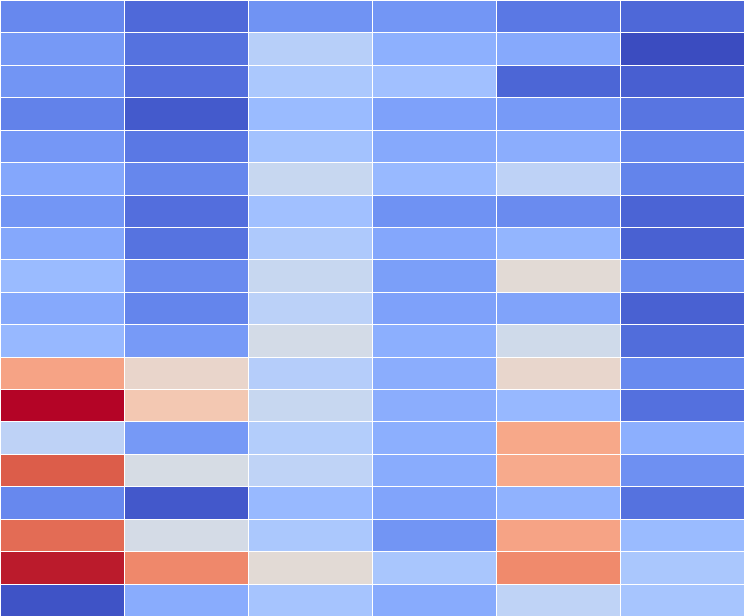};
\end{axis}

\end{tikzpicture}

%% file: figure_scripts/citizenships.tex
\begin{tikzpicture}

\begin{axis}[
  width=\linewidth,height=0.5\linewidth,
axis y line*=left,
axis x line*=bottom,
tick align=outside,
tick pos=left,
unbounded coords=jump,
x grid style={darkgray176},
xmin=-0.65, xmax=14.5,
xtick style={color=black},
xtick={0,1,2,3,4,5,6,7,8,9,10,11,12,13,14},
xticklabel style={rotate=90.0},
xticklabels={
  United States,
  United Kingdom,
  France,
  Germany,
  Italy,
  China,
  Russia,
  India,
  Austria-Hungary,
  Spain,
  Netherlands,
  Japan,
  Greece,
  Egypt,
  Iran
},
y grid style={darkgray176},
ymin=0, ymax=134.4,
ytick style={color=black},
ticklabel style={font=\scriptsize},
]
\draw[draw=none,fill=darksalmon233149163] (axis cs:-0.4,0) rectangle (axis cs:0.4,128);
\draw[draw=none,fill=darksalmon228148112] (axis cs:0.6,0) rectangle (axis cs:1.4,59);
\draw[draw=none,fill=peru19815275] (axis cs:1.6,0) rectangle (axis cs:2.4,44);
\draw[draw=none,fill=peru17115771] (axis cs:2.6,0) rectangle (axis cs:3.4,37);
\draw[draw=none,fill=yellowgreen14416469] (axis cs:3.6,0) rectangle (axis cs:4.4,21);
\draw[draw=none,fill=olivedrab9517471] (axis cs:4.6,0) rectangle (axis cs:5.4,17);
\draw[draw=none,fill=mediumseagreen73174130] (axis cs:5.6,0) rectangle (axis cs:6.4,15);
\draw[draw=none,fill=cadetblue74172155] (axis cs:6.6,0) rectangle (axis cs:7.4,11);
\draw[draw=none,fill=cadetblue75170172] (axis cs:7.6,0) rectangle (axis cs:8.4,9);
\draw[draw=none,fill=cadetblue80171194] (axis cs:8.6,0) rectangle (axis cs:9.4,9);
\draw[draw=none,fill=cornflowerblue108173226] (axis cs:9.6,0) rectangle (axis cs:10.4,8);
\draw[draw=none,fill=lightsteelblue173171235] (axis cs:10.6,0) rectangle (axis cs:11.4,7);
\draw[draw=none,fill=plum206155232] (axis cs:11.6,0) rectangle (axis cs:12.4,7);
\draw[draw=none,fill=violet230139219] (axis cs:12.6,0) rectangle (axis cs:13.4,7);
\draw[draw=none,fill=plum232145191] (axis cs:13.6,0) rectangle (axis cs:14.4,5);

\end{axis}

\end{tikzpicture}

%% file: figure_scripts/languages.tex
\begin{tikzpicture}

\begin{axis}[
width=\linewidth,height=0.5\linewidth,
axis y line*=left,
axis x line*=bottom,
tick align=outside,
tick pos=left,
unbounded coords=jump,
x grid style={darkgray176},
xmin=-0.65, xmax=14.5,
xtick style={color=black},
xtick={0,1,2,3,4,5,6,7,8,9,10,11,12,13,14},
xticklabel style={rotate=90.0},
xticklabels={
  English,
  French,
  German,
  Latin,
  Spanish,
  Italian,
  Chinese,
  Russian,
  Arabic,
  Ancient Greek,
  Dutch,
  Portuguese,
  Greek,
  Japanese,
  Polish
},
y grid style={darkgray176},
ymin=0, ymax=303.45,
ytick style={color=black},
ticklabel style={font=\scriptsize},
]
\draw[draw=none,fill=darksalmon233149163] (axis cs:-0.4,0) rectangle (axis cs:0.4,289);
\draw[draw=none,fill=darksalmon228148112] (axis cs:0.6,0) rectangle (axis cs:1.4,103);
\draw[draw=none,fill=peru19815275] (axis cs:1.6,0) rectangle (axis cs:2.4,76);
\draw[draw=none,fill=peru17115771] (axis cs:2.6,0) rectangle (axis cs:3.4,54);
\draw[draw=none,fill=yellowgreen14416469] (axis cs:3.6,0) rectangle (axis cs:4.4,45);
\draw[draw=none,fill=olivedrab9517471] (axis cs:4.6,0) rectangle (axis cs:5.4,33);
\draw[draw=none,fill=mediumseagreen73174130] (axis cs:5.6,0) rectangle (axis cs:6.4,32);
\draw[draw=none,fill=cadetblue74172155] (axis cs:6.6,0) rectangle (axis cs:7.4,28);
\draw[draw=none,fill=cadetblue75170172] (axis cs:7.6,0) rectangle (axis cs:8.4,23);
\draw[draw=none,fill=cadetblue80171194] (axis cs:8.6,0) rectangle (axis cs:9.4,21);
\draw[draw=none,fill=cornflowerblue108173226] (axis cs:9.6,0) rectangle (axis cs:10.4,13);
\draw[draw=none,fill=lightsteelblue173171235] (axis cs:10.6,0) rectangle (axis cs:11.4,10);
\draw[draw=none,fill=plum206155232] (axis cs:11.6,0) rectangle (axis cs:12.4,9);
\draw[draw=none,fill=violet230139219] (axis cs:12.6,0) rectangle (axis cs:13.4,9);
\draw[draw=none,fill=plum232145191] (axis cs:13.6,0) rectangle (axis cs:14.4,8);
\end{axis}

\end{tikzpicture}

%% file: figure_scripts/wiki_factscores.tex
\begin{tikzpicture}

\definecolor{brown1926061}{RGB}{192,60,61}
\definecolor{burlywood238187136}{RGB}{238,187,136}
\definecolor{darkgray176}{RGB}{176,176,176}
\definecolor{darkseagreen159212148}{RGB}{159,212,148}
\definecolor{darkslategray61}{RGB}{61,61,61}
\definecolor{dimgray1319183}{RGB}{131,91,83}
\definecolor{gray127}{RGB}{127,127,127}
\definecolor{lightpink241164163}{RGB}{241,164,163}
\definecolor{lightseagreen45171184}{RGB}{45,171,184}
\definecolor{lightsteelblue181200224}{RGB}{181,200,224}
\definecolor{mediumpurple147113178}{RGB}{147,113,178}
\definecolor{orchid213132188}{RGB}{213,132,188}
\definecolor{peru22412844}{RGB}{224,128,44}
\definecolor{pink238190211}{RGB}{238,190,211}
\definecolor{rosybrown190160154}{RGB}{190,160,154}
\definecolor{seagreen5814558}{RGB}{58,145,58}
\definecolor{silver196180208}{RGB}{196,180,208}
\definecolor{silver199}{RGB}{199,199,199}
\definecolor{steelblue49115161}{RGB}{49,115,161}
\definecolor{tan209209150}{RGB}{209,209,150}
\definecolor{yellowgreen16816953}{RGB}{168,169,53}

\begin{axis}[
width=0.7\linewidth,height=0.33\linewidth,
axis y line*=left,
axis x line*=bottom,
tick align=outside,
tick pos=left,
x grid style={darkgray176},
xlabel={Language},
xmin=-0.65, xmax=18.5,
xtick style={color=black},
xtick={0,1,2,3,4,5,6,7,8,9,10,11,12,13,14,15,16,17,18},
xticklabels={\vphantom{h}en,ja,zh,\vphantom{h}es,\vphantom{h}fr,pl,\vphantom{h}vi,\vphantom{h}tr,fa,ko,\vphantom{h}ar,hu,th,hi,bn,\vphantom{h}ms,\vphantom{h}ta,\vphantom{h}sw,jv},
y grid style={darkgray176},
ylabel={Factscores},
ymin=0.242268041237113, ymax=1.0360824742268,
ytick style={color=black},
ticklabel style={font=\scriptsize},
]
\path [draw=darkslategray61, fill=steelblue49115161, semithick]
(axis cs:-0.4,0.945454545454545)
--(axis cs:0.4,0.945454545454545)
--(axis cs:0.4,0.983870967741935)
--(axis cs:-0.4,0.983870967741935)
--(axis cs:-0.4,0.945454545454545)
--cycle;
\path [draw=darkslategray61, fill=lightsteelblue181200224, semithick]
(axis cs:0.6,0.652962833099579)
--(axis cs:1.4,0.652962833099579)
--(axis cs:1.4,0.857142857142857)
--(axis cs:0.6,0.857142857142857)
--(axis cs:0.6,0.652962833099579)
--cycle;
\path [draw=darkslategray61, fill=peru22412844, semithick]
(axis cs:1.6,0.666030534351145)
--(axis cs:2.4,0.666030534351145)
--(axis cs:2.4,0.842617908407382)
--(axis cs:1.6,0.842617908407382)
--(axis cs:1.6,0.666030534351145)
--cycle;
\path [draw=darkslategray61, fill=burlywood238187136, semithick]
(axis cs:2.6,0.646008403361345)
--(axis cs:3.4,0.646008403361345)
--(axis cs:3.4,0.840571305841924)
--(axis cs:2.6,0.840571305841924)
--(axis cs:2.6,0.646008403361345)
--cycle;
\path [draw=darkslategray61, fill=seagreen5814558, semithick]
(axis cs:3.6,0.638876603272888)
--(axis cs:4.4,0.638876603272888)
--(axis cs:4.4,0.811202506854681)
--(axis cs:3.6,0.811202506854681)
--(axis cs:3.6,0.638876603272888)
--cycle;
\path [draw=darkslategray61, fill=darkseagreen159212148, semithick]
(axis cs:4.6,0.622626582278481)
--(axis cs:5.4,0.622626582278481)
--(axis cs:5.4,0.833333333333333)
--(axis cs:4.6,0.833333333333333)
--(axis cs:4.6,0.622626582278481)
--cycle;
\path [draw=darkslategray61, fill=brown1926061, semithick]
(axis cs:5.6,0.707107843137255)
--(axis cs:6.4,0.707107843137255)
--(axis cs:6.4,0.901436956081765)
--(axis cs:5.6,0.901436956081765)
--(axis cs:5.6,0.707107843137255)
--cycle;
\path [draw=darkslategray61, fill=lightpink241164163, semithick]
(axis cs:6.6,0.645644796380091)
--(axis cs:7.4,0.645644796380091)
--(axis cs:7.4,0.876605868358446)
--(axis cs:6.6,0.876605868358446)
--(axis cs:6.6,0.645644796380091)
--cycle;
\path [draw=darkslategray61, fill=mediumpurple147113178, semithick]
(axis cs:7.6,0.678261916225325)
--(axis cs:8.4,0.678261916225325)
--(axis cs:8.4,0.901456310679612)
--(axis cs:7.6,0.901456310679612)
--(axis cs:7.6,0.678261916225325)
--cycle;
\path [draw=darkslategray61, fill=silver196180208, semithick]
(axis cs:8.6,0.639508196721312)
--(axis cs:9.4,0.639508196721312)
--(axis cs:9.4,0.865909090909091)
--(axis cs:8.6,0.865909090909091)
--(axis cs:8.6,0.639508196721312)
--cycle;
\path [draw=darkslategray61, fill=dimgray1319183, semithick]
(axis cs:9.6,0.666666666666667)
--(axis cs:10.4,0.666666666666667)
--(axis cs:10.4,0.888211382113821)
--(axis cs:9.6,0.888211382113821)
--(axis cs:9.6,0.666666666666667)
--cycle;
\path [draw=darkslategray61, fill=rosybrown190160154, semithick]
(axis cs:10.6,0.666666666666667)
--(axis cs:11.4,0.666666666666667)
--(axis cs:11.4,0.865909090909091)
--(axis cs:10.6,0.865909090909091)
--(axis cs:10.6,0.666666666666667)
--cycle;
\path [draw=darkslategray61, fill=orchid213132188, semithick]
(axis cs:11.6,0.688806731259561)
--(axis cs:12.4,0.688806731259561)
--(axis cs:12.4,0.883624031007752)
--(axis cs:11.6,0.883624031007752)
--(axis cs:11.6,0.688806731259561)
--cycle;
\path [draw=darkslategray61, fill=pink238190211, semithick]
(axis cs:12.6,0.684210526315789)
--(axis cs:13.4,0.684210526315789)
--(axis cs:13.4,0.91304347826087)
--(axis cs:12.6,0.91304347826087)
--(axis cs:12.6,0.684210526315789)
--cycle;
\path [draw=darkslategray61, fill=gray127, semithick]
(axis cs:13.6,0.715401785714286)
--(axis cs:14.4,0.715401785714286)
--(axis cs:14.4,0.901630434782609)
--(axis cs:13.6,0.901630434782609)
--(axis cs:13.6,0.715401785714286)
--cycle;
\path [draw=darkslategray61, fill=silver199, semithick]
(axis cs:14.6,0.705882352941177)
--(axis cs:15.4,0.705882352941177)
--(axis cs:15.4,0.906976744186046)
--(axis cs:14.6,0.906976744186046)
--(axis cs:14.6,0.705882352941177)
--cycle;
\path [draw=darkslategray61, fill=yellowgreen16816953, semithick]
(axis cs:15.6,0.721153846153846)
--(axis cs:16.4,0.721153846153846)
--(axis cs:16.4,0.9)
--(axis cs:15.6,0.9)
--(axis cs:15.6,0.721153846153846)
--cycle;
\path [draw=darkslategray61, fill=tan209209150, semithick]
(axis cs:16.6,0.686677631578947)
--(axis cs:17.4,0.686677631578947)
--(axis cs:17.4,0.909090909090909)
--(axis cs:16.6,0.909090909090909)
--(axis cs:16.6,0.686677631578947)
--cycle;
\path [draw=darkslategray61, fill=lightseagreen45171184, semithick]
(axis cs:17.6,0.609577922077922)
--(axis cs:18.4,0.609577922077922)
--(axis cs:18.4,0.831214689265537)
--(axis cs:17.6,0.831214689265537)
--(axis cs:17.6,0.609577922077922)
--cycle;
\addplot [semithick, darkslategray61]
table {%
0 0.945454545454545
0 0.88785046728972
};
\addplot [semithick, darkslategray61]
table {%
0 0.983870967741935
0 1
};
\addplot [semithick, darkslategray61]
table {%
-0.2 0.88785046728972
0.2 0.88785046728972
};
\addplot [semithick, darkslategray61]
table {%
-0.2 1
0.2 1
};
\addplot [semithick, darkslategray61]
table {%
1 0.652962833099579
1 0.35
};
\addplot [semithick, darkslategray61]
table {%
1 0.857142857142857
1 1
};
\addplot [semithick, darkslategray61]
table {%
0.8 0.35
1.2 0.35
};
\addplot [semithick, darkslategray61]
table {%
0.8 1
1.2 1
};
\addplot [semithick, darkslategray61]
table {%
2 0.666030534351145
2 0.427385892116183
};
\addplot [semithick, darkslategray61]
table {%
2 0.842617908407382
2 1
};
\addplot [semithick, darkslategray61]
table {%
1.8 0.427385892116183
2.2 0.427385892116183
};
\addplot [semithick, darkslategray61]
table {%
1.8 1
2.2 1
};
\addplot [semithick, darkslategray61]
table {%
3 0.646008403361345
3 0.371794871794872
};
\addplot [semithick, darkslategray61]
table {%
3 0.840571305841924
3 1
};
\addplot [semithick, darkslategray61]
table {%
2.8 0.371794871794872
3.2 0.371794871794872
};
\addplot [semithick, darkslategray61]
table {%
2.8 1
3.2 1
};
\addplot [semithick, darkslategray61]
table {%
4 0.638876603272888
4 0.385714285714286
};
\addplot [semithick, darkslategray61]
table {%
4 0.811202506854681
4 1
};
\addplot [semithick, darkslategray61]
table {%
3.8 0.385714285714286
4.2 0.385714285714286
};
\addplot [semithick, darkslategray61]
table {%
3.8 1
4.2 1
};
\addplot [semithick, darkslategray61]
table {%
5 0.622626582278481
5 0.333333333333333
};
\addplot [semithick, darkslategray61]
table {%
5 0.833333333333333
5 1
};
\addplot [semithick, darkslategray61]
table {%
4.8 0.333333333333333
5.2 0.333333333333333
};
\addplot [semithick, darkslategray61]
table {%
4.8 1
5.2 1
};
\addplot [semithick, darkslategray61]
table {%
6 0.707107843137255
6 0.447272727272727
};
\addplot [semithick, darkslategray61]
table {%
6 0.901436956081765
6 1
};
\addplot [semithick, darkslategray61]
table {%
5.8 0.447272727272727
6.2 0.447272727272727
};
\addplot [semithick, darkslategray61]
table {%
5.8 1
6.2 1
};
\addplot [semithick, darkslategray61]
table {%
7 0.645644796380091
7 0.313609467455621
};
\addplot [semithick, darkslategray61]
table {%
7 0.876605868358446
7 1
};
\addplot [semithick, darkslategray61]
table {%
6.8 0.313609467455621
7.2 0.313609467455621
};
\addplot [semithick, darkslategray61]
table {%
6.8 1
7.2 1
};
\addplot [semithick, darkslategray61]
table {%
8 0.678261916225325
8 0.380952380952381
};
\addplot [semithick, darkslategray61]
table {%
8 0.901456310679612
8 1
};
\addplot [semithick, darkslategray61]
table {%
7.8 0.380952380952381
8.2 0.380952380952381
};
\addplot [semithick, darkslategray61]
table {%
7.8 1
8.2 1
};
\addplot [semithick, darkslategray61]
table {%
9 0.639508196721312
9 0.326086956521739
};
\addplot [semithick, darkslategray61]
table {%
9 0.865909090909091
9 1
};
\addplot [semithick, darkslategray61]
table {%
8.8 0.326086956521739
9.2 0.326086956521739
};
\addplot [semithick, darkslategray61]
table {%
8.8 1
9.2 1
};
\addplot [semithick, darkslategray61]
table {%
10 0.666666666666667
10 0.39568345323741
};
\addplot [semithick, darkslategray61]
table {%
10 0.888211382113821
10 1
};
\addplot [semithick, darkslategray61]
table {%
9.8 0.39568345323741
10.2 0.39568345323741
};
\addplot [semithick, darkslategray61]
table {%
9.8 1
10.2 1
};
\addplot [semithick, darkslategray61]
table {%
11 0.666666666666667
11 0.368421052631579
};
\addplot [semithick, darkslategray61]
table {%
11 0.865909090909091
11 1
};
\addplot [semithick, darkslategray61]
table {%
10.8 0.368421052631579
11.2 0.368421052631579
};
\addplot [semithick, darkslategray61]
table {%
10.8 1
11.2 1
};
\addplot [semithick, darkslategray61]
table {%
12 0.688806731259561
12 0.4
};
\addplot [semithick, darkslategray61]
table {%
12 0.883624031007752
12 1
};
\addplot [semithick, darkslategray61]
table {%
11.8 0.4
12.2 0.4
};
\addplot [semithick, darkslategray61]
table {%
11.8 1
12.2 1
};
\addplot [semithick, darkslategray61]
table {%
13 0.684210526315789
13 0.384615384615385
};
\addplot [semithick, darkslategray61]
table {%
13 0.91304347826087
13 1
};
\addplot [semithick, darkslategray61]
table {%
12.8 0.384615384615385
13.2 0.384615384615385
};
\addplot [semithick, darkslategray61]
table {%
12.8 1
13.2 1
};
\addplot [semithick, darkslategray61]
table {%
14 0.715401785714286
14 0.447513812154696
};
\addplot [semithick, darkslategray61]
table {%
14 0.901630434782609
14 1
};
\addplot [semithick, darkslategray61]
table {%
13.8 0.447513812154696
14.2 0.447513812154696
};
\addplot [semithick, darkslategray61]
table {%
13.8 1
14.2 1
};
\addplot [semithick, darkslategray61]
table {%
15 0.705882352941177
15 0.409090909090909
};
\addplot [semithick, darkslategray61]
table {%
15 0.906976744186046
15 1
};
\addplot [semithick, darkslategray61]
table {%
14.8 0.409090909090909
15.2 0.409090909090909
};
\addplot [semithick, darkslategray61]
table {%
14.8 1
15.2 1
};
\addplot [semithick, darkslategray61]
table {%
16 0.721153846153846
16 0.455357142857143
};
\addplot [semithick, darkslategray61]
table {%
16 0.9
16 1
};
\addplot [semithick, darkslategray61]
table {%
15.8 0.455357142857143
16.2 0.455357142857143
};
\addplot [semithick, darkslategray61]
table {%
15.8 1
16.2 1
};
\addplot [semithick, darkslategray61]
table {%
17 0.686677631578947
17 0.361702127659574
};
\addplot [semithick, darkslategray61]
table {%
17 0.909090909090909
17 1
};
\addplot [semithick, darkslategray61]
table {%
16.8 0.361702127659574
17.2 0.361702127659574
};
\addplot [semithick, darkslategray61]
table {%
16.8 1
17.2 1
};
\addplot [semithick, darkslategray61]
table {%
18 0.609577922077922
18 0.278350515463918
};
\addplot [semithick, darkslategray61]
table {%
18 0.831214689265537
18 1
};
\addplot [semithick, darkslategray61]
table {%
17.8 0.278350515463918
18.2 0.278350515463918
};
\addplot [semithick, darkslategray61]
table {%
17.8 1
18.2 1
};
\addplot [semithick, darkslategray61]
table {%
-0.4 0.966666666666667
0.4 0.966666666666667
};
\addplot [semithick, darkslategray61]
table {%
0.6 0.757166257166257
1.4 0.757166257166257
};
\addplot [semithick, darkslategray61]
table {%
1.6 0.752049180327869
2.4 0.752049180327869
};
\addplot [semithick, darkslategray61]
table {%
2.6 0.739565217391304
3.4 0.739565217391304
};
\addplot [semithick, darkslategray61]
table {%
3.6 0.739130434782609
4.4 0.739130434782609
};
\addplot [semithick, darkslategray61]
table {%
4.6 0.729281767955801
5.4 0.729281767955801
};
\addplot [semithick, darkslategray61]
table {%
5.6 0.805309734513274
6.4 0.805309734513274
};
\addplot [semithick, darkslategray61]
table {%
6.6 0.771929824561403
7.4 0.771929824561403
};
\addplot [semithick, darkslategray61]
table {%
7.6 0.803738317757009
8.4 0.803738317757009
};
\addplot [semithick, darkslategray61]
table {%
8.6 0.758620689655172
9.4 0.758620689655172
};
\addplot [semithick, darkslategray61]
table {%
9.6 0.790088826554465
10.4 0.790088826554465
};
\addplot [semithick, darkslategray61]
table {%
10.6 0.769230769230769
11.4 0.769230769230769
};
\addplot [semithick, darkslategray61]
table {%
11.6 0.8
12.4 0.8
};
\addplot [semithick, darkslategray61]
table {%
12.6 0.8
13.4 0.8
};
\addplot [semithick, darkslategray61]
table {%
13.6 0.817744755244755
14.4 0.817744755244755
};
\addplot [semithick, darkslategray61]
table {%
14.6 0.818181818181818
15.4 0.818181818181818
};
\addplot [semithick, darkslategray61]
table {%
15.6 0.813953488372093
16.4 0.813953488372093
};
\addplot [semithick, darkslategray61]
table {%
16.6 0.812169312169312
17.4 0.812169312169312
};
\addplot [semithick, darkslategray61]
table {%
17.6 0.733333333333333
18.4 0.733333333333333
};
\end{axis}

\end{tikzpicture}

%% file: figure_scripts/fs_exp_multi.tex
\begin{tikzpicture}

\definecolor{darkslategray38}{RGB}{38,38,38}
\definecolor{lightgray204}{RGB}{204,204,204}

\begin{axis}[width=\linewidth, height=\linewidth,
axis line style={lightgray204},
colorbar,
colormap={mymap}{[1pt]
  rgb(0pt)=(0.705673158,0.01555616,0.150232812);
  rgb(1pt)=(0.752534934,0.157246067,0.184115123);
  rgb(2pt)=(0.795631745,0.24128379,0.220525627);
  rgb(3pt)=(0.834620542,0.312874446,0.259301199);
  rgb(4pt)=(0.869186849,0.378313092,0.300267182);
  rgb(5pt)=(0.89904617,0.439559467,0.343229596);
  rgb(6pt)=(0.923944917,0.49730856,0.387970225);
  rgb(7pt)=(0.943660866,0.551750968,0.434243684);
  rgb(8pt)=(0.958003065,0.602842431,0.481775914);
  rgb(9pt)=(0.966811177,0.650421156,0.530263762);
  rgb(10pt)=(0.969954137,0.694266682,0.579375448);
  rgb(11pt)=(0.96732803,0.734132809,0.628751763);
  rgb(12pt)=(0.958852946,0.769767752,0.678007945);
  rgb(13pt)=(0.944468518,0.800927443,0.726736146);
  rgb(14pt)=(0.924127593,0.827384882,0.774508472);
  rgb(15pt)=(0.897787179,0.848937047,0.820880546);
  rgb(16pt)=(0.865395197,0.86541021,0.865395561);
  rgb(17pt)=(0.833466556,0.860207984,0.901068838);
  rgb(18pt)=(0.798691636,0.849786142,0.931688648);
  rgb(19pt)=(0.761464949,0.834302879,0.956945269);
  rgb(20pt)=(0.722193294,0.813952739,0.976574709);
  rgb(21pt)=(0.681291281,0.788964712,0.990363227);
  rgb(22pt)=(0.639176211,0.759599947,0.998151185);
  rgb(23pt)=(0.596262162,0.726149107,0.999836203);
  rgb(24pt)=(0.552953156,0.688929332,0.995375608);
  rgb(25pt)=(0.509635204,0.648280772,0.98478814);
  rgb(26pt)=(0.46666708,0.604562568,0.968154911);
  rgb(27pt)=(0.424369608,0.558148092,0.945619588);
  rgb(28pt)=(0.38301334,0.50941904,0.917387822);
  rgb(29pt)=(0.342804478,0.458757618,0.883725899);
  rgb(30pt)=(0.30386891,0.406535296,0.84495867);
  rgb(31pt)=(0.26623388,0.353094838,0.801466763);
  rgb(32pt)=(0.2298057,0.298717966,0.753683153)
},
colorbar/width=2.5mm,
point meta max=0.746794606901669,
point meta min=0.107728723818126,
tick align=outside,
tick pos=left,
x grid style={lightgray204},
xlabel={Model},
xlabel style={yshift=-7ex},
xmin=0, xmax=6,
xtick style={color=darkslategray38},
xtick={0.5,1.5,2.5,3.5,4.5,5.5},
xticklabels={Aya8, Aya35, LLaMA8, LLaMA70, Qwen7, Qwen72},
xticklabel style={rotate=45.0},
y dir=reverse,
y grid style={lightgray204},
ylabel={Language},
ymin=0, ymax=19,
ytick style={color=darkslategray38},
ytick={0.5,1.5,2.5,3.5,4.5,5.5,6.5,7.5,8.5,9.5,10.5,11.5,12.5,13.5,14.5,15.5,16.5,17.5,18.5},
yticklabels={en, ja, zh, es, fr, pl, vi, tr, fa, ko, ar, hu, th, hi, bn, ms, ta, sw, jv},
ticklabel style={font=\scriptsize}
]
\addplot graphics [includegraphics cmd=\pgfimage,xmin=0, xmax=6, ymin=19, ymax=0] {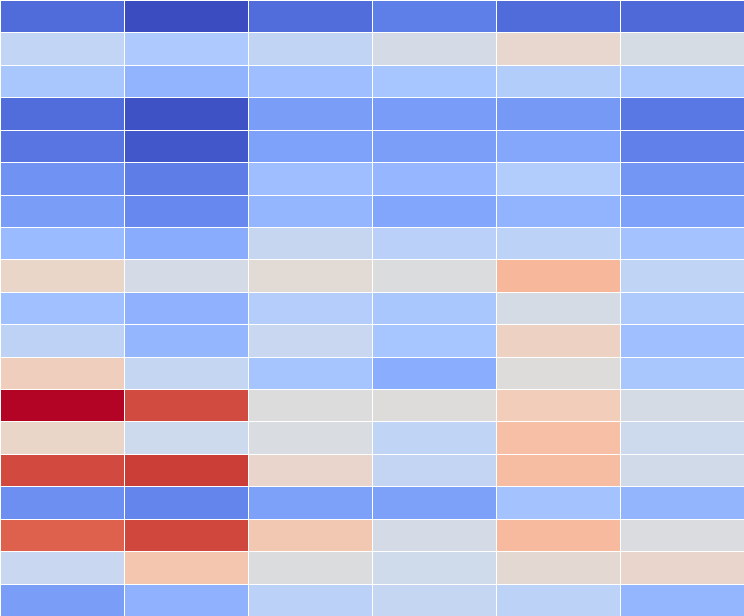};
\end{axis}

\end{tikzpicture}

%% file: figure_scripts/fs_exp_trans.tex
\begin{tikzpicture}

\definecolor{darkslategray38}{RGB}{38,38,38}
\definecolor{lightgray204}{RGB}{204,204,204}

\begin{axis}[width=\linewidth, height=\linewidth,
axis line style={lightgray204},
colorbar,
colormap={mymap}{[1pt]
  rgb(0pt)=(0.705673158,0.01555616,0.150232812);
  rgb(1pt)=(0.752534934,0.157246067,0.184115123);
  rgb(2pt)=(0.795631745,0.24128379,0.220525627);
  rgb(3pt)=(0.834620542,0.312874446,0.259301199);
  rgb(4pt)=(0.869186849,0.378313092,0.300267182);
  rgb(5pt)=(0.89904617,0.439559467,0.343229596);
  rgb(6pt)=(0.923944917,0.49730856,0.387970225);
  rgb(7pt)=(0.943660866,0.551750968,0.434243684);
  rgb(8pt)=(0.958003065,0.602842431,0.481775914);
  rgb(9pt)=(0.966811177,0.650421156,0.530263762);
  rgb(10pt)=(0.969954137,0.694266682,0.579375448);
  rgb(11pt)=(0.96732803,0.734132809,0.628751763);
  rgb(12pt)=(0.958852946,0.769767752,0.678007945);
  rgb(13pt)=(0.944468518,0.800927443,0.726736146);
  rgb(14pt)=(0.924127593,0.827384882,0.774508472);
  rgb(15pt)=(0.897787179,0.848937047,0.820880546);
  rgb(16pt)=(0.865395197,0.86541021,0.865395561);
  rgb(17pt)=(0.833466556,0.860207984,0.901068838);
  rgb(18pt)=(0.798691636,0.849786142,0.931688648);
  rgb(19pt)=(0.761464949,0.834302879,0.956945269);
  rgb(20pt)=(0.722193294,0.813952739,0.976574709);
  rgb(21pt)=(0.681291281,0.788964712,0.990363227);
  rgb(22pt)=(0.639176211,0.759599947,0.998151185);
  rgb(23pt)=(0.596262162,0.726149107,0.999836203);
  rgb(24pt)=(0.552953156,0.688929332,0.995375608);
  rgb(25pt)=(0.509635204,0.648280772,0.98478814);
  rgb(26pt)=(0.46666708,0.604562568,0.968154911);
  rgb(27pt)=(0.424369608,0.558148092,0.945619588);
  rgb(28pt)=(0.38301334,0.50941904,0.917387822);
  rgb(29pt)=(0.342804478,0.458757618,0.883725899);
  rgb(30pt)=(0.30386891,0.406535296,0.84495867);
  rgb(31pt)=(0.26623388,0.353094838,0.801466763);
  rgb(32pt)=(0.2298057,0.298717966,0.753683153)
},
colorbar/width=2.5mm,
point meta max=0.812875925101466,
point meta min=0.209059763437455,
tick align=outside,
tick pos=left,
x grid style={lightgray204},
xlabel={Model},
xlabel style={yshift=-7ex},
xmin=0, xmax=6,
xtick style={color=darkslategray38},
xtick={0.5,1.5,2.5,3.5,4.5,5.5},
xticklabels={Aya8, Aya35, LLaMA8, LLaMA70, Qwen7, Qwen72},
xticklabel style={rotate=45.0},
y dir=reverse,
y grid style={lightgray204},
ylabel={Language},
ymin=0, ymax=19,
ytick style={color=darkslategray38},
ytick={0.5,1.5,2.5,3.5,4.5,5.5,6.5,7.5,8.5,9.5,10.5,11.5,12.5,13.5,14.5,15.5,16.5,17.5,18.5},
yticklabels={en, ja, zh, es, fr, pl, vi, tr, fa, ko, ar, hu, th, hi, bn, ms, ta, sw, jv},
ticklabel style={font=\scriptsize},
]
\addplot graphics [includegraphics cmd=\pgfimage,xmin=0, xmax=6, ymin=19, ymax=0] {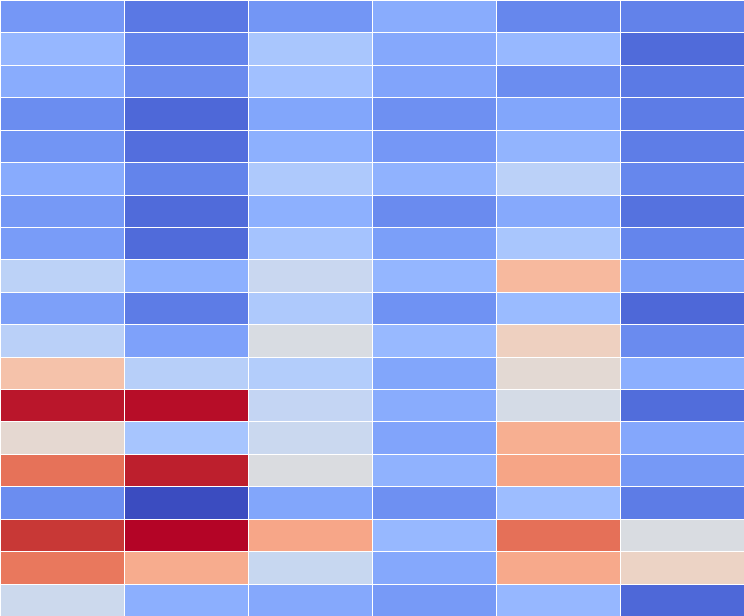};
\end{axis}

\end{tikzpicture}